\documentclass[lettersize,journal]{IEEEtran}
\usepackage{amsmath,amsfonts}
\usepackage{algorithmic}
\usepackage{algorithm}
\usepackage{array}
\usepackage[caption=false,font=normalsize,labelfont=sf,textfont=sf]{subfig}
\usepackage{textcomp}
\usepackage{stfloats}
\usepackage{url}
\usepackage{verbatim}
\usepackage{graphicx}
\usepackage{cite}
\usepackage[utf8]{inputenc}
\usepackage{microtype}
\usepackage{inconsolata}
\usepackage{epstopdf}
\usepackage{epsfig}
\usepackage{microtype}
\usepackage{amssymb}
\usepackage{booktabs}
\usepackage{hyperref}
\usepackage{cleveref}
\usepackage{tabularx}
\usepackage{multirow}
\usepackage{makecell}
\usepackage[table]{xcolor}
\usepackage[switch]{lineno}

\definecolor{upcolor}{gray}{0.4} 

\begin{document}

\title{Dep-LLM: Training-Free Depression Diagnosis via Evidence-Guided Structured Multi-factor with Reliable LLM Reasoning}

\author{Yiqing Lyu, Xianbing Zhao, Buzhou Tang, Ronghuan Jiang
\thanks{
Yiqing Lyu is with School of Computer Science and Technology, Harbin Institute of Technology, Shenzhen, Guangdong, China (e-mail: kosmischer@stu.hit.edu.cn).}
\thanks{
Xianbing Zhao is with School of Artificial Intelligence and Computer Science,
Jiangnan University, Wuxi, China, Harbin Institute of Technology, Shenzhen, Guangdong, China, and Guangdong Provincial Key Laboratory of Intelligent Information Processing (e-mail: zhaoxianbing\_hitsz@163.com).}
\thanks{
Buzhou Tang (Corresponding Author) is with School of Computer Science and Technology, Harbin Institute of Technology, Shenzhen, Guangdong, China, Pengcheng Laboratory, Guangdong Provincial Key Laboratory of Intelligent Information Processing (e-mail: tangbuzhou@hit.edu.cn).}
\thanks{Ronghuan Jiang (Corresponding Author) is with Chinese People’s Liberation Army General Hospital, Beijing, China (e-mail: jangrh55@126.com).}
\thanks{This study is partially supported by National Key R\&D Program of China (2023YFC3502900), National Natural Science Foundation of China (62276082), Shenzhen Science and Technology Research and Development Fund (KJZD20240903102802003), Shenzhen Science and Technology Research and Development Fund for Sustainable Development Project (GXWD20231128103819001, 20230706140548006) and Guangdong Provincial Key Laboratory Grant (2023B1212060076).}
}

\markboth{Journal of \LaTeX\ Class Files,~Vol.~14, No.~8, August~2021}%
{Shell \MakeLowercase{\textit{et al.}}: A Sample Article Using IEEEtran.cls for IEEE Journals}


\maketitle

\begin{abstract}
Automatic Depression Detection (ADD) from clinical interviews is a pivotal task in computational mental health, yet it remains challenging due to two critical obstacles: 1) difficulty in modeling complex but sparsely distributed depression clues within lengthy, multi-topic clinical interviews, leading to superficial and unreliable reasoning; 2) scarcity of labeled data due to clinical privacy, together with high cost of training and fine-tuning, limiting the deployment of supervised ADD systems. To jointly address these challenges, we propose Dep-LLM, a training-free framework that mirrors the step-by-step reasoning of clinical psychiatrists and operates entirely on frozen off-the-shelf foundation LLMs. Dep-LLM comprises three stages. First, a Chain-of-Thought (CoT) Depression Multi-factor Analysis module structurally decomposes the long dialogue into five clinically aligned themes and produces evidence-grounded rationales, effectively handling long-context dependencies. Second, we introduce Confidence Analysis and Modulation module that quantifies the epistemic reliability from token-level entropy of each rationale and applies an intra-label and inter-theme modulation that amplifies trustworthy signals while suppressing uncertain ones without extra training. Third, a Collaborative Multi-factor Prediction module dynamically integrates multi-factor signals weighted by confidence into the final diagnosis. Extensive experiments on the DAIC-WOZ and E-DAIC datasets demonstrate the effectiveness and generalizability of Dep-LLM: it surpasses zero-shot baseline on nearly all 21 foundation LLMs across 9 metrics such as accuracy, macro F1 and weighted-average F1, and further outperforms state-of-the-art supervised domain-specific LLMs as well as the latest closed-source commercial LLMs, while requiring no extra training.
\end{abstract}


\section{Introduction}\label{sec:intro}

\begin{figure}[t]
\centering
\begin{center}
\centerline{\includegraphics[width=\linewidth]{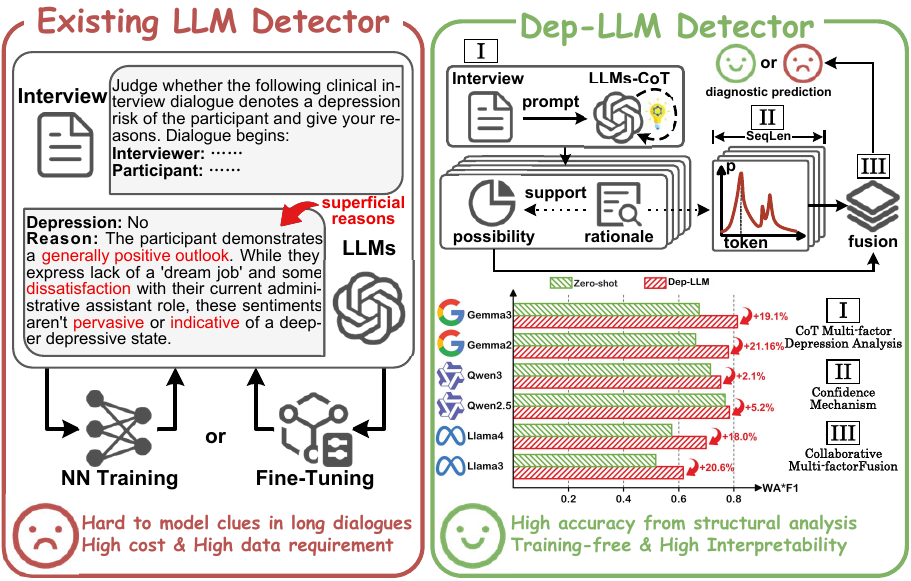}}
\caption{
Dep-LLM decomposes and analyzes dialogues via structural multi-factor schema and verifies their reliability by confidence mechanism. Without extra training, Dep-LLM outperforms zero-shot settings on a range of foundation LLMs. }
\label{fig:Intro}
\end{center}
\end{figure}

\IEEEPARstart{M}ental disorders, particularly depression, have become a major global health challenge. According to recent statistics from WHO, depression affects millions of people worldwide, standing as a leading cause of disability and contributing significantly to the global burden of disease~\cite{who2024world, loweimi2025zero}. The authoritative psychiatric literature DSM-5~\cite{apa2013diagnostic} establishes the gold standard for depression diagnosis, which is operationalized through Structured Clinical Interview for DSM-5 (SCID), where clinicians assess a patient's psychiatric state via complex dialogue interaction. Psychiatric works~\cite{brodey2018rapid, shankman2018reliability} also propose a synthetic methodology for SCID design by structurally incorporating themes such as family relationship, work satisfaction, medical history and so on. While this synthetic approach is clinically validated, the manual administration and analysis of such interviews are resource-intensive and difficult to scale to meet the growing demand for mental health services~\cite{teferra2024leveraging}. Consequently, Automatic Depression Detection (ADD) from clinical interview transcripts has attracted extensive attention, aiming to assist clinicians by objectively identifying depressive risks from natural language~\cite{zhang2025speecht, ge2025survey}.

Early ADD research focused on supervised deep learning over interview transcripts~\cite{rinaldi2020predicting, Jung2024Hique, zogan2022explainable}, which suffers from limited interpretability and heavy reliance on labeled clinical data~\cite{zhang2025mitigating}. The rise of Large Language Models (LLMs) has shifted ADD toward generative reasoning, domain-adaptive pre-training~\cite{ji-etal-2022-mentalbert}, instruction fine-tuning~\cite{xu2024mental, yang2023mentalllama}, retrieval-augmented generation~\cite{zhang2025explainable, zhang2025speecht}, and multi-agent pipelines~\cite{hu2025agentmental, zhou2024intermind}. Despite the progress, ADD remains challenging due to two bottlenecks that persist across both supervised and LLM-based methods: difficulty in modeling sparse depression clues within long clinical interview dialogues, and the practical barriers of data scarcity and high training cost.

\textbf{Challenge 1: Difficulty in modeling depression clues within long context.} Clinical interviews are inherently lengthy and cover multiple themes, in which depression clues are complex but sparsely distributed, and intertwined with content irrelevant to depression diagnosis~\cite{Yao2024Depression, Jung2024Hique} (e.g., greetings and transitional sentences). This complex semantic nature makes it difficult for off-the-shelf LLMs to capture the subtle symptoms associated with depression. As illustrated in Figure~\ref{fig:Intro}, zero-shot LLMs tend to collapse the entire dialogue into a holistic judgment and produce superficial reasons such as ``a generally positive outlook'', overlooking specific clues that are clinically relevant~\cite{zhang2025explainable, ringeval2019avec}. Moreover, even when LLMs produce detailed rationales, these may be locally plausible yet clinically unreliable due to medical hallucination~\cite{kim2025medical, Asgari2025framework}. Therefore, a structured analysis schema well-aligned with the clinical standard (e.g., SCID) is needed to decompose these interview dialogues with multiple factors into evidence-grounded rationales~\cite{zhao-etal-2025-predicting, SARMA2026integrating, Lee2026interpretable}, accompanied by a reliability verification mechanism to distinguish trustworthy rationales from uncertain ones. 

\textbf{Challenge 2: Data scarcity and high training cost.} Supervised methods~\cite{zogan2022explainable, mao2025enhancing} and fine-tuned LLMs~\cite{ji-etal-2022-mentalbert, yang2023mentalllama, xu2024mental} rely heavily on large-scale labeled clinical data, which is scarce and hard to access due to privacy and ethical concerns. Without such data, these models are difficult to optimize and deploy in domain-specific clinical settings. Furthermore, training or fine-tuning LLMs calls for expensive computational resources, creating a prohibitive barrier for clinical institutions with limited budgets~\cite{Zhou2025large, miao2024integrating}. Consequently, training-free methods that leverage frozen foundation LLMs without extra data or training overhead are urgently needed, while their performance must remain competitive against supervised models.

To jointly address these two challenges, we propose \textbf{Dep-LLM}, a novel training-free framework for ADD that mirrors the step-by-step reasoning process of clinical psychiatrists. As illustrated in Figure~\ref{fig:Intro} and Figure~\ref{fig:model}, Dep-LLM is a three-stage pipeline operating entirely on frozen off-the-shelf foundation LLMs: 1) The \emph{CoT Depression Multi-factor Analysis} module uses Chain-of-Thought prompting~\cite{wei2022chain, teng2025enhancing} to structurally decompose the long dialogue into five SCID-aligned themes~\cite{ringeval2019avec, zhao-etal-2025-predicting} (family relationship, work satisfaction, mental state, medical history, overall evaluation) and extract evidence-based rationales over a fine-grained possibility space, effectively handling semantic dependencies in long context; 2) The \emph{Semantic Confidence Analysis and Modulation} module quantifies the epistemic reliability of each rationale from token-level entropy and applies an intra-label and inter-theme contrastive modulation that amplifies reliable signals while suppressing uncertain ones~\cite{farquhar2024detecting, nguyen2025beyond, PennyDimri2025Measuring}; 3) The \emph{Collaborative Multi-factor Prediction} module dynamically integrates these confidence-weighted signals into a diagnostic decision. The entire pipeline introduces no learnable parameters and requires no labeled clinical data, making Dep-LLM accessible in deployment under the constrained data and computational budgets typical of real-world clinical settings.

Extensive experiments on the widely used DAIC-WOZ~\cite{daic-gratch2014distress} and E-DAIC~\cite{Devault2014Simsensei} datasets demonstrate the effectiveness of Dep-LLM. Against zero-shot baselines, Dep-LLM yields significant improvements on nearly all tested 21 foundation LLMs (Llama-2/3/4, Qwen-2.5/3/3.5, Gemma-2/3 families, with parameters ranging from 4B to 17B) and further outperforms representative supervised domain-specific LLMs (e.g., MentalBERT, MentaLLaMA, BioMistral, Meditron) as well as the latest closed-source commercial LLMs (GPT-5.5, Gemini-3.1-Pro, Claude-Opus-4.6, Grok-4.3, DeepSeek-V4) across most metrics. Our main contributions are three-fold:

\begin{itemize}
    \item We propose \textbf{Dep-LLM}, a novel depression detection framework addressing the challenge of reasoning in long SCID dialogues via introducing structured multi-factor analysis schema and LLM reliability verification.
    \item We implement Dep-LLM framework under a fully training-free setting, eliminating the overhead for expensive model training and the need for scarce data while retaining strong performance in clinical scenarios.
    \item We realize the Dep-LLM architecture, integrating CoT reasoning, Semantic Confidence analysis, and Collaborative Multi-factor fusion, which jointly reinforces the interpretability and rationality of the automatic diagnosis.
\end{itemize}

\section{Related Work}

\subsection{Automatic Depression Detection}

The methodology for Automatic Depression Detection (ADD) has followed a clear trajectory from supervised deep learning and multimodal fusion to the current paradigm of generative reasoning with Large Language Models (LLMs).

Early ADD research leveraged neural networks like CNNs and RNNs for sequential dialogue feature modeling~\cite{gong2017topic, rinaldi2020predicting}, alongside frameworks like TFN~\cite{zadeh2017tensor}, MulT~\cite{tsai2019multimodal}, and MISA~\cite{hazarika2020misa} to perform multimodal fusion. HAN~\cite{mallol2019hierarchical} further refined this via hierarchical attention. Recent advances prioritize complex fusion: DepMSTAT~\cite{tao2024depmstat} and TTFNet~\cite{chen2025ttfnet} utilize spatio-temporal and frequency-domain networks, while DepMamba~\cite{ye2024depmamba} employs state space models for efficiency. Methods like MMPF~\cite{yang2024mmpf} and WavFace~\cite{flores2025wavface} focus on signal filtration and alignment. Structural approaches like SEGA~\cite{chen2024depression} and HiQuE~\cite{Jung2024Hique} focus on reconstructing semantic structure via embedding networks. However, these supervised models universally suffer from low interpretability and high dependency for training data~\cite{zhang2025mitigating}.

The advancement of LLMs has shifted the focus toward generative reasoning. Initial efforts leveraged pre-trained models, such as MentalBERT~\cite{ji-etal-2022-mentalbert}, or instruction fine-tuning models, such as MentalAlpaca~\cite{xu2024mental}, to align models with mental health tasks. More recent supervised domain-specific LLMs like PsycoLLM~\cite{jinpeng2025psycollm} and DepressLLM~\cite{moon2025depressllm} further inject psychological knowledge and interpretable confidence into the backbone. Beyond text, multimodal LLMs fuse linguistic clues with acoustic and facial signals for interview-based assessment~\cite{Sadeghi2024harnessing, Li2025large}. Meanwhile, agentic frameworks decompose diagnosis across collaborative roles, including doctor-patient-family interaction~\cite{zhou2024intermind}, multi-agent guided interviewing~\cite{bi2025magi, hu2025agentmental}, knowledge-guided psychiatric reasoning~\cite{wu2026wisemind}, and realistic patient simulation~\cite{liu2025eeyore}. To address the hallucination and the lack of clinical grounding of off-the-shelf LLMs~\cite{ohse2024zeroshot}, Retrieval-Augmented Generation (RAG) methods such as SpeechT-RAG~\cite{zhang2025speecht} and RED~\cite{zhang2025explainable} ground diagnosis on external evidence. Apart from that, Chain-of-Thought (CoT) frameworks such as Doris~\cite{lan2025depression}, EMDRC~\cite{zheng2025towards}, and emotion-to-reasoning prompting~\cite{teng2025enhancing} structure the reasoning process around psychiatric criteria. Yet most of these methods either lack an internal mechanism to verify the reliability of their own reasoning, or intensively depend on scarce labeled data and computational resources for fine-tuning, retrieval, or simulation.

Dep-LLM addresses these mentioned gaps via elaborating a training-free framework mirroring clinical assessment through CoT-driven structured multi-factor analysis and employing semantic confidence mechanism to mathematically reinforce reasoning reliability, ensuring both interpretability and accuracy without training overhead.

\section{Method} \label{sec:method}
\begin{figure*}[ht]
\centering
\begin{center}
\centerline{\includegraphics[width=\textwidth]{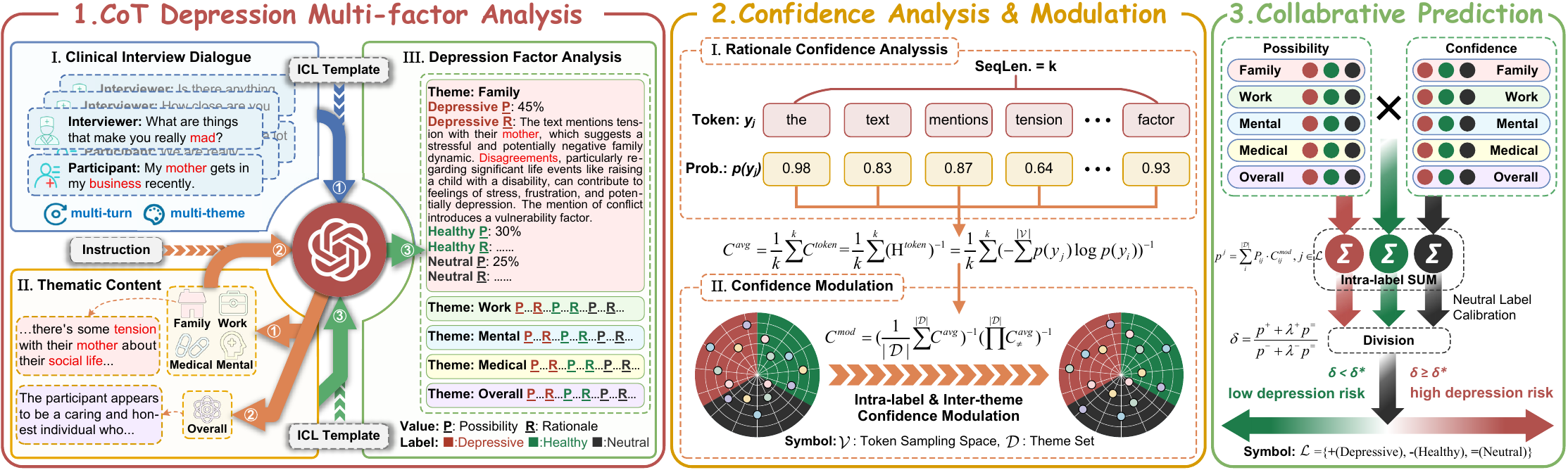}}
\caption{
Schematic illustration of the proposed Dep-LLM framework with three components. The CoT Depression Multi-factor Analysis applies CoT techniques to prompt LLMs to assess possibilities and rationales for each theme and each label. Confidence Analysis \& Modulation introduces confidence mechanism to evaluate the reliability of those rationales. Collaborative Prediction incorporates possibility, rationale and confidence across multi-factor.
}
\label{fig:model}
\end{center}
\end{figure*}

In this section, we elaborate the methodological architecture of our Dep-LLM, a fully training-free framework that mirrors the step-by-step reasoning process of clinical psychiatrists. To jointly address the two bottlenecks of inaccurate clue reasoning in long SCID-style dialogues and the prohibitive cost of supervised adaptation which are identified in Section~\ref{sec:intro}, Dep-LLM is designed as a three-stage pipeline that operates entirely on the frozen off-the-shelf foundation LLMs, requiring no task-specific fine-tuning, no labeled depression corpus, and no auxiliary trainable parameters. As illustrated in Figure~\ref{fig:model}, the pipeline 1)~structurally decomposes the raw interview dialogues into a multi-theme, multi-label evidence space through CoT-driven prompting; 2)~quantifies the epistemic reliability from token-level entropy of every generated rationale, with intra-label and inter-theme modulation; 3)~synthesizes a diagnostic prediction based on a collaborative fusion strategy weighted by confidence. Notably, all of the three stages use the same frozen LLM only at inference time. Therefore the entire Dep-LLM framework can be deployed under constrained data and resource of clinical settings without any extra training overhead.

\subsection{Problem Formulation}\label{sec:problem}

Given clinical interview dialogue transcript $\mathcal S$, our goal is to assess the participant's depressive state by extracting and analyzing multi-factor evidence from $\mathcal S$ in a training-free manner. To this end, we define collaborative multi-factor space spanning two orthogonal dimensions: multi-theme $\mathcal D$ aligned with SCID standard~\cite{brodey2018rapid, shankman2018reliability}, and multi-label $\mathcal L$ for fine-grained diagnostic analysis:

\begin{gather}
    \mathcal{D} = \{\text{family,work,mental,medical,overall}\},\\
    \mathcal{L} = \{\text{+(depressive),-(healthy),=(neutral)}\},
\end{gather}

For each theme $i\in\mathcal D$ and each potential label $j\in\mathcal L$, Dep-LLM derives a possibility score $P_{ij}$, and an evidence-grounded rationale $R_{ij}$, and  a semantic confidence $C_{ij}$ that measures the reliability of $R_{ij}$. The final diagnosis $\hat y$ is obtained by dynamically fusing these factors in $\mathcal{D}\times\mathcal{L}$ weighted by their confidences. Throughout the pipeline, the foundation LLM is invoked only as a frozen decoder, which is the key property that enables Dep-LLM to remain training-free.

\subsection{CoT Depression Multi-factor Analysis}\label{sec:cot}

Standard zero-shot prompting tends to collapse the long multi-turn dialogue into a single holistic judgment, which often yields vague and superficial rationales and overlooks sparsely distributed depression clues~\cite{Yao2024Depression, ringeval2019avec}. To mitigate this, the first stage of Dep-LLM employs a Chain-of-Thought (CoT) scheme that decomposes the reasoning process into three sequential prompting steps, all executed on the same frozen LLM without parameter update.

\subsubsection{Step 1: Thematic Content Extraction}

For a multi-turn clinical interview dialogue $\mathcal{S}$ and a frozen large language model $\mathrm{LLM}(\cdot)$, we employ an in-context learning (ICL) template $X_1$ (provided in supplemental file) to prompt the LLM to extract the narrative thematic content of the four atomic themes $\mathcal D^\circ_{\{\text{family,work,mental,medical}\}}$:
\begin{equation}\label{eq:t1}
    \mathcal{T}^{\circ}=\bigl\{\,t_{i}\mid t_{i}=\mathrm{LLM}(\mathcal{S},X_{1}),\,i\in\mathcal{D}^{\circ}\bigr\},
\end{equation}
where $t_i$ retains only the clinically relevant content for theme $i$ and filters out trivial chitchat such as greetings and transitional sentences. This decomposition explicitly realigns the unstructured dialogue with the structured multi-theme schema corresponding to SCID.

\subsubsection{Step2: Integrative Overall Synthesis}

We then apply a second instruction $X_2$ that prompts the LLM to integrate the four atomic themes into an \textit{overall} narrative summary, capturing global clues (e.g., cross-theme emotional consistency) that a single theme cannot express in isolation:
\begin{equation}\label{eq:t2}
    \mathcal{T}=\mathcal{T}^{\circ}\cup\bigl\{t_{\text{overall}}\mid t_{\text{overall}}=\mathrm{LLM}(\mathcal{T}^{\circ},X_{2})\bigr\}.
\end{equation}
The overall theme contributes integrative evidence rather than a redundant factor, which is verified through experiments in ablation study (Section~\ref{sec:ablation}).

\subsubsection{Step3: Depression Multi-factor Analysis}

Finally, we employ a third ICL template $X_3$ to prompt the LLM to perform a fine-grained assessment over the full multi-factor space $\mathcal{D}\times\mathcal{L}$. For every $\text{(theme,label)}$ pair, this step produces a possibility score $P_{ij}$ and an explicit justifying rationale $R_{ij}$:
\begin{equation}\label{eq:t3}
    (P_{ij},R_{ij})=\mathrm{LLM}(\mathcal{T},X_{3}),\quad i\in\mathcal{D},\,j\in\mathcal{L}.
\end{equation}
The three-step decomposition ensures that the LLM never has to simultaneously perform topic filtering, cross-theme integration, and a label-conditioned reasoning in a single forward pass, thereby substantially mitigating the superficial summary observed in the zero-shot baseline (Figure~\ref{fig:Intro}). Instead, the multi-factor rationales are strictly evidence-grounded and directed by CoT-driven structured analysis schema.

\subsection{Semantic Confidence Analysis and Modulation}\label{sec:conf}

The rationales generated by stage~\ref{sec:cot} are linguistically fluent but not all equally trustworthy given that training-free LLMs are prone to producing locally plausible yet clinically unreliable explanations~\cite{kim2025medical, wu2024uncertainty}. Instead of verifying these rationales via supervised fine-tuning which intensively calls for resource and labeled data, we address this by exploiting the epistemic reliability, a quantity that exists inherently during generation which can be calculated using the model's token-level predictive distribution.

\subsubsection{Theoretical Grounding}

The motivation behind our confidence mechanism is straightforward: when an LLM is genuinely confident in what it is saying, its probability distribution of next token should be sharply peaked on a small set of candidates; when it is fabricating content, by contrast, the distribution tends to flatten because no single continuation is strongly supported by the model's internal knowledge. There are works that theoretically support the motivation. Generally,~\cite{kadavath2022language}~showed that the predictive entropy of an LLM is inversely correlated with the quality (i.e., confidence) of its generation, establishing entropy as a robust proxy for the model's self-knowledge. Specifically in medical fields where hallucinations are high stakes,~\cite{manakul2023selfcheckgpt, farquhar2024detecting} verified that high-entropy (i.e., low-confidence) states are a reliable signal of non-factual generation, and~\cite{wu2024uncertainty} further demonstrated that entropy-based uncertainty estimation is critical for distinguishing trustworthy medical advice from hazardous hallucinations. Mathematically, we define the rationale confidence $C$ as the reciprocal of the Shannon entropy $H$ of the LLM's token distribution, aligning with the geometric framework proposed by~\cite{phillips2025geometric} which demonstrates low uncertainty corresponds to a tighter concentration of semantic embeddings, indicating a stable and trustworthy evidence.

\subsubsection{Token-Level Entropy and Rationale-Level Confidence}

Formally, for a generated token $y_k$ at step $k$, the token-level entropy is:
\begin{equation}\label{eq:Htoken}
    H^{\text{token}}(y_{k})=-\sum_{v\in\mathcal{V}}p(v\mid y_{<k},x)\log_{2}p(v\mid y_{<k},x),
\end{equation}
where $\mathcal{V}$ is the token sampling space and $p(v\mid \cdot)$ is the probability of candidate token $v$. A low $H^{\text{token}}$ indicates that the model is decisive in its selection, reflecting internalized knowledge rather than hallucinated medical details. We then define the token confidence as the reciprocal of the token entropy:
\begin{equation}\label{eq:Ctoken}
    C^{\text{token}}(y_{k})=\frac{1}{H^{\text{token}}(y_{k})+\epsilon},
\end{equation}
where $\epsilon$ is a small constant for numerical stability. Then we average it along the sequence to obtain the rationale-level average confidence:
\begin{equation}\label{eq:cavg}
    C^{\text{avg}}(y)=\frac{1}{|y|}\sum_{k=1}^{|y|}C^{\text{token}}(y_{k}).
\end{equation}
In practice, we apply Equation~\ref{eq:cavg} to every rationale produced in Equation~\ref{eq:t3}:
\begin{equation}\label{eq:cavg_ij}
C^{\text{avg}}_{ij}=C^{\text{avg}}(R_{ij}),\quad i\in\mathcal{D},\,j\in\mathcal{L},
\end{equation}
yielding a $|\mathcal{D}\times\mathcal{L}|$ raw average confidence matrix that measures the extent to which each rationale is internally consistent with the LLM's own probabilistic beliefs. Notably, this process is totally training-free and introduces no learnable parameters since the entropy is derived directly from the same forward pass that produces the rationale.

\subsubsection{Intra-Label and Inter-Theme Confidence Modulation}

Raw Confidence scores may exhibit high variance across different themes due to varying text lengths and inherent semantic complexities, so a naive employment of $C^{\text{avg}}_{ij}$ would cause scale-driven artifacts dominating the fusion. To address this bias, we apply a contrastive modulation based on the principle that a factor should weigh more when its own rationale is confident and its competing themes are uncertain. Concretely, we first compute a harmonic-mean normalizer:
\begin{equation}\label{eq:alpha}
    \alpha^{\text{norm}}_{j}=\frac{1}{\frac{1}{|\mathcal{D}|}\sum_{i\in\mathcal{D}}\frac{1}{C^{\text{avg}}_{ij}}},\quad j\in\mathcal{L},
\end{equation}
and then derive the modulated confidence as:
\begin{equation}\label{eq:cmod}
    C^{\text{mod}}_{ij}=\frac{\alpha^{\text{norm}}_{j}}{\prod_{i'\in\mathcal{D},\,i'\neq i}C^{\text{avg}}_{i'j}},\quad j\in\mathcal{L}.
\end{equation}
By construction, for theme $i\in\mathcal{D}$, $C^\text{mod}_{ij}$ is monotonically increasing in its own raw confidence $C^\text{avg}_{ij}$ and decreasing in the confidences of the competing themes $i'\in\mathcal D,i'\neq i$ under the same label $j\in\mathcal L$. This effectively amplifies the signal of the most reliable diagnostic factor while suppressing uncertain ones, which is a behaviour quantitatively verified later in case study in Section~\ref{sec:case}.

\subsection{Collaborative Multi-factor Prediction}\label{sec:pred}
In the final stage, Dep-LLM synthesizes the multi-factor assessments into a single diagnostic decision via a confidence-weighted fusion. Notably, this stage involves no learned weights, where all coefficients are either derived analytically from the modulated confidence (Equation~\ref{eq:cmod}) or fixed by design and exposed to the clinician as transparent hyperparameters.

\subsubsection{Confidence-Weighted Aggregation}
First, for each label $j\in\mathcal L$, the support score $p^{j}$ is computed by fusing the possibilities with the modulated confidences of the five themes:
\begin{equation}\label{eq:pj}
p^{j}=\sum_{i\in\mathcal{D}}P_{ij}\cdot C^{\text{mod}}_{ij},\quad j\in\mathcal{L}_{\{+,-,=\}}.
\end{equation}
In this formulation, possibility of each theme contributes to the final decision in proportion to how reliably the LLM justified it, rather than flat summation based on the possibility itself only.

\subsubsection{Collaborative Prediction}
Afterwards, we define depressive coefficient $\delta$ to reflect the predictive risk of depression. While depressive evidence $p^+$ and healthy evidence $p^-$ vote for their own label, the neutral evidence $p^=$ is intrinsically ambiguous yet should not be discarded directly. Therefore we introduce neutral label calibrators $\lambda^+, \lambda^-$, explicitly incorporating the neutral evidence:
\begin{equation}\label{eq:delta}
\delta=\frac{p^{+}+\lambda^{+}\cdot p^{=}}{p^{-}+\lambda^{-}\cdot p^{=}},
\end{equation}
where a higher $\delta$ indicates a higher predictive risk of depression. The final prediction is obtained by filtering $\delta$ through the decision threshold $\delta^*$:
\begin{equation}\label{eq:y}
    \hat{y}=\begin{cases}
        \text{depressive}\;(+), & \text{if }\delta\geq\delta^{*},\\
        \text{healthy}\;(-), & \text{otherwise}.
    \end{cases}
\end{equation}

\subsubsection{Default Configuration}

In our standard configuration, we assign $\lambda^+=\lambda^-=1$ and $\delta^*=1$, assuming balanced sensitivity towards the two diagnostic poles. As $\lambda^\pm$ and $\delta^*$ are decision-side hyperparameters rather than learnable weights, they can be calibrated on a small held-out validation set or directly adjusted by clinicians to reflect domain-specific risk preference. We systematically explore this property in the hyperparameter sensitivity analysis in Section~\ref{sec:hyp}.

\begin{table*}[!ht]
\begin{center}
\renewcommand{\arraystretch}{1.00}
\caption{Performance Comparison on DAIC-WOZ between zero-shot baseline and Dep-LLM. Ma* refers to macro metrics, WA* refers to weighted-average metrics., F1-Depr. and F1-Heal. refer to per-class binary metrics. For readability, gray shade marks the performance of Dep-LLM and Bold marks the performance of the better one.}
\label{tab:Comparison}
\newcommand{\up}[1]{\;{\scriptsize\color{upcolor}{\textbf{$\uparrow$}#1}}} 
\begin{tabular}{ l | c | >{\centering\arraybackslash}p{1.06cm} | >{\centering\arraybackslash}p{1.06cm} | >{\centering\arraybackslash}p{1.06cm} | >{\centering\arraybackslash}p{1.06cm} | >{\centering\arraybackslash}p{1.06cm} | >{\centering\arraybackslash}p{1.06cm} | >{\centering\arraybackslash}p{1.06cm} | >{\centering\arraybackslash}p{1.06cm} | >{\centering\arraybackslash}p{1.06cm} }
\toprule
\textbf{Foundation LLM} & \textbf{Method} & \textbf{Acc.} & \textbf{Ma*Prec.} & \textbf{Ma*Rec.} & \textbf{Ma*F1.} & \textbf{WA*Prec.} & \textbf{WA*Rec.} & \textbf{WA*F1.} & \textbf{F1-Depr.} & \textbf{F1-Heal.}\\

\midrule
\multirow{2}{*}{Llama2-7B-Base}
& Zero-shot & 0.468 & 0.516 & 0.518 & 0.464 & 0.600 & 0.468 & 0.483 & 0.419 & 0.510\\
& \cellcolor{gray!15}Dep-LLM & \cellcolor{gray!15}\textbf{0.574} & \cellcolor{gray!15}\textbf{0.579} & \cellcolor{gray!15}\textbf{0.594} & \cellcolor{gray!15}\textbf{0.558} & \cellcolor{gray!15}\textbf{0.661} & \cellcolor{gray!15}\textbf{0.574} & \cellcolor{gray!15}\textbf{0.592} & \cellcolor{gray!15}\textbf{0.474} & \cellcolor{gray!15}\textbf{0.643}\\

\multirow{2}{*}{Llama2-7B-Chat}
& Zero-shot & 0.574 & 0.562 & 0.574 & 0.550 & 0.642 & 0.574 & 0.592 & 0.444 & 0.655\\
& \cellcolor{gray!15}Dep-LLM & \cellcolor{gray!15}\textbf{0.638} & \cellcolor{gray!15}\textbf{0.618} & \cellcolor{gray!15}\textbf{0.640} & \cellcolor{gray!15}\textbf{0.613} & \cellcolor{gray!15}\textbf{0.695} & \cellcolor{gray!15}\textbf{0.638} & \cellcolor{gray!15}\textbf{0.653} & \cellcolor{gray!15}\textbf{0.514} & \cellcolor{gray!15}\textbf{0.712}\\

\multirow{2}{*}{Llama2-13B-Base}
& Zero-shot & 0.510 & 0.506 & 0.508 & 0.487 & 0.588 & 0.511 & 0.531 & 0.378 & 0.596\\
& \cellcolor{gray!15}Dep-LLM & \cellcolor{gray!15}\textbf{0.574} & \cellcolor{gray!15}\textbf{0.545} & \cellcolor{gray!15}\textbf{0.553} & \cellcolor{gray!15}\textbf{0.539} & \cellcolor{gray!15}\textbf{0.624} & \cellcolor{gray!15}\textbf{0.574} & \cellcolor{gray!15}\textbf{0.591} & \cellcolor{gray!15}\textbf{0.412} & \cellcolor{gray!15}\textbf{0.667}\\

\multirow{2}{*}{Llama2-13B-Chat}
& Zero-shot & \textbf{0.553} & 0.549 & 0.558 & \textbf{0.532} & 0.630 & \textbf{0.553} & \textbf{0.572} & 0.432 & \textbf{0.632}\\
& \cellcolor{gray!15}Dep-LLM & \cellcolor{gray!15}0.532 & \cellcolor{gray!15}\textbf{0.595} & \cellcolor{gray!15}\textbf{0.605} & \cellcolor{gray!15}0.530 & \cellcolor{gray!15}\textbf{0.687} & \cellcolor{gray!15}0.532 & \cellcolor{gray!15}0.542 & \cellcolor{gray!15}\textbf{0.500} & \cellcolor{gray!15}0.560\\

\multirow{2}{*}{Llama3-8B-Base}
& Zero-shot & 0.532 & 0.554 & 0.521 & 0.450 & 0.638 & 0.532 & 0.550 & 0.450 & 0.593\\
& \cellcolor{gray!15}Dep-LLM & \cellcolor{gray!15}\textbf{0.681} & \cellcolor{gray!15}\textbf{0.677} & \cellcolor{gray!15}\textbf{0.711} & \cellcolor{gray!15}\textbf{0.666} & \cellcolor{gray!15}\textbf{0.757} & \cellcolor{gray!15}\textbf{0.681} & \cellcolor{gray!15}\textbf{0.694} & \cellcolor{gray!15}\textbf{0.595} & \cellcolor{gray!15}\textbf{0.737}\\

\multirow{2}{*}{Llama3-8B-Instruct}
& Zero-shot & 0.532 & 0.519 & 0.523 & 0.505 & 0.600 & 0.532 & 0.552 & 0.389 & 0.621\\
& \cellcolor{gray!15}Dep-LLM & \cellcolor{gray!15}\textbf{0.660} & \cellcolor{gray!15}\textbf{0.647} & \cellcolor{gray!15}\textbf{0.675} & \cellcolor{gray!15}\textbf{0.640} & \cellcolor{gray!15}\textbf{0.725} & \cellcolor{gray!15}\textbf{0.660} & \cellcolor{gray!15}\textbf{0.674} & \cellcolor{gray!15}\textbf{0.556} & \cellcolor{gray!15}\textbf{0.724}\\

\multirow{2}{*}{Llama4-17B-Base}
& Zero-shot & 0.596 & 0.609 & 0.630 & 0.584 & 0.694 & 0.596 & 0.612 & 0.513 & 0.655\\
& \cellcolor{gray!15}Dep-LLM & \cellcolor{gray!15}\textbf{0.681} & \cellcolor{gray!15}\textbf{0.635} & \cellcolor{gray!15}\textbf{0.649} & \cellcolor{gray!15}\textbf{0.639} & \cellcolor{gray!15}\textbf{0.702} & \cellcolor{gray!15}\textbf{0.681} & \cellcolor{gray!15}\textbf{0.689} & \cellcolor{gray!15}\textbf{0.516} & \cellcolor{gray!15}\textbf{0.762}\\

\multirow{2}{*}{Llama4-17B-Instruct}
& Zero-shot & 0.617 & 0.589 & 0.604 & 0.585 & 0.665 & 0.617 & 0.631 & 0.471 & 0.700\\
& \cellcolor{gray!15}Dep-LLM & \cellcolor{gray!15}\textbf{0.744} & \cellcolor{gray!15}\textbf{0.719} & \cellcolor{gray!15}\textbf{0.756} & \cellcolor{gray!15}\textbf{0.724} & \cellcolor{gray!15}\textbf{0.788} & \cellcolor{gray!15}\textbf{0.745} & \cellcolor{gray!15}\textbf{0.754} & \cellcolor{gray!15}\textbf{0.647} & \cellcolor{gray!15}\textbf{0.800}\\

\midrule
\multirow{2}{*}{Qwen2.5-7B-Base}
& Zero-shot & 0.660 & 0.619 & 0.634 & 0.621 & 0.689 & 0.660 & 0.670 & 0.500 & 0.742\\
& \cellcolor{gray!15}Dep-LLM & \cellcolor{gray!15}\textbf{0.681} & \cellcolor{gray!15}\textbf{0.635} & \cellcolor{gray!15}\textbf{0.649} & \cellcolor{gray!15}\textbf{0.639} & \cellcolor{gray!15}\textbf{0.702} & \cellcolor{gray!15}\textbf{0.681} & \cellcolor{gray!15}\textbf{0.689} & \cellcolor{gray!15}\textbf{0.516} & \cellcolor{gray!15}\textbf{0.762}\\

\multirow{2}{*}{Qwen2.5-7B-Instruct}
& Zero-shot & 0.702 & 0.690 & 0.726 & 0.685 & 0.767 & 0.702 & 0.715 & 0.611 & 0.759\\
& \cellcolor{gray!15}Dep-LLM & \cellcolor{gray!15}\textbf{0.787} & \cellcolor{gray!15}\textbf{0.763} & \cellcolor{gray!15}\textbf{0.807} & \cellcolor{gray!15}\textbf{0.770} & \cellcolor{gray!15}\textbf{0.829} & \cellcolor{gray!15}\textbf{0.787} & \cellcolor{gray!15}\textbf{0.795} & \cellcolor{gray!15}\textbf{0.706} & \cellcolor{gray!15}\textbf{0.833}\\

\multirow{2}{*}{Qwen2.5-14B-Base}
& Zero-shot & 0.617 & 0.558 & 0.563 & 0.559 & 0.633 & 0.617 & 0.624 & 0.400 & 0.719\\
& \cellcolor{gray!15}Dep-LLM & \cellcolor{gray!15}\textbf{0.660} & \cellcolor{gray!15}\textbf{0.619} & \cellcolor{gray!15}\textbf{0.634} & \cellcolor{gray!15}\textbf{0.621} & \cellcolor{gray!15}\textbf{0.689} & \cellcolor{gray!15}\textbf{0.660} & \cellcolor{gray!15}\textbf{0.670} & \cellcolor{gray!15}\textbf{0.500} & \cellcolor{gray!15}\textbf{0.742}\\

\multirow{2}{*}{Qwen2.5-14B-Instruct}
& Zero-shot & \textbf{0.809} & 0.773 & \textbf{0.761} & \textbf{0.766} & \textbf{0.805} & \textbf{0.809} & \textbf{0.806} & \textbf{0.667} & 0.866\\
& \cellcolor{gray!15}Dep-LLM & \cellcolor{gray!15}\textbf{0.809} & \cellcolor{gray!15}\textbf{0.780} & \cellcolor{gray!15}0.740 & \cellcolor{gray!15}0.755 & \cellcolor{gray!15}0.802 & \cellcolor{gray!15}\textbf{0.809} & \cellcolor{gray!15}0.801 & \cellcolor{gray!15}0.640 & \cellcolor{gray!15}\textbf{0.870}\\

\multirow{2}{*}{Qwen3-4B-Base}
& Zero-shot & \textbf{0.681} & 0.635 & 0.649 & 0.639 & 0.702 & \textbf{0.681} & 0.689 & 0.516 & \textbf{0.762}\\
& \cellcolor{gray!15}Dep-LLM & \cellcolor{gray!15}\textbf{0.681} & \cellcolor{gray!15}\textbf{0.661} & \cellcolor{gray!15}\textbf{0.690} & \cellcolor{gray!15}\textbf{0.659} & \cellcolor{gray!15}\textbf{0.736} & \cellcolor{gray!15}\textbf{0.681} & \cellcolor{gray!15}\textbf{0.694} & \cellcolor{gray!15}\textbf{0.571} & \cellcolor{gray!15}0.746\\

\multirow{2}{*}{Qwen3-8B-Base}
& Zero-shot & 0.766 & 0.722 & 0.731 & 0.726 & 0.771 & 0.766 & 0.768 & 0.621 & 0.831\\
& \cellcolor{gray!15}Dep-LLM & \cellcolor{gray!15}\textbf{0.787} & \cellcolor{gray!15}\textbf{0.754} & \cellcolor{gray!15}\textbf{0.787} & \cellcolor{gray!15}\textbf{0.763} & \cellcolor{gray!15}\textbf{0.812} & \cellcolor{gray!15}\textbf{0.787} & \cellcolor{gray!15}\textbf{0.794} & \cellcolor{gray!15}\textbf{0.688} & \cellcolor{gray!15}\textbf{0.839}\\

\multirow{2}{*}{Qwen3-14B-Base}
& Zero-shot & \textbf{0.702} & 0.644 & 0.644 & 0.644 & 0.702 & \textbf{0.702} & 0.702 & 0.500 & \textbf{0.788}\\
& \cellcolor{gray!15}Dep-LLM & \cellcolor{gray!15}\textbf{0.702} & \cellcolor{gray!15}\textbf{0.664} & \cellcolor{gray!15}\textbf{0.685} & \cellcolor{gray!15}\textbf{0.668} & \cellcolor{gray!15}\textbf{0.730} & \cellcolor{gray!15}\textbf{0.702} & \cellcolor{gray!15}\textbf{0.711} & \cellcolor{gray!15}\textbf{0.563} & \cellcolor{gray!15}0.774\\

\multirow{2}{*}{Qwen3.5-4B-Base}
& Zero-shot & 0.723 & 0.681 & 0.700 & 0.687 & 0.743 & 0.723 & 0.730 & 0.581 & 0.794\\
& \cellcolor{gray!15}Dep-LLM & \cellcolor{gray!15}\textbf{0.745} & \cellcolor{gray!15}\textbf{0.701} & \cellcolor{gray!15}\textbf{0.715} & \cellcolor{gray!15}\textbf{0.706} & \cellcolor{gray!15}\textbf{0.756} & \cellcolor{gray!15}\textbf{0.745} & \cellcolor{gray!15}\textbf{0.749} & \cellcolor{gray!15}\textbf{0.600} & \cellcolor{gray!15}\textbf{0.813}\\

\multirow{2}{*}{Qwen3.5-9B-Base}
& Zero-shot & 0.723 & 0.666 & 0.659 & 0.662 & 0.718 & 0.723 & 0.720 & 0.518 & 0.806\\
& \cellcolor{gray!15}Dep-LLM & \cellcolor{gray!15}\textbf{0.787} & \cellcolor{gray!15}\textbf{0.754} & \cellcolor{gray!15}\textbf{0.787} & \cellcolor{gray!15}\textbf{0.763} & \cellcolor{gray!15}\textbf{0.812} & \cellcolor{gray!15}\textbf{0.787} & \cellcolor{gray!15}\textbf{0.794} & \cellcolor{gray!15}\textbf{0.688} & \cellcolor{gray!15}\textbf{0.839}\\

\midrule
\multirow{2}{*}{Gemma2-9B-Base}
& Zero-shot & 0.660 & 0.647 & 0.675 & 0.640 & 0.725 & 0.660 & 0.674 & 0.556 & 0.724\\
& \cellcolor{gray!15}Dep-LLM & \cellcolor{gray!15}\textbf{0.745} & \cellcolor{gray!15}\textbf{0.732} & \cellcolor{gray!15}\textbf{0.777} & \cellcolor{gray!15}\textbf{0.730} & \cellcolor{gray!15}\textbf{0.808} & \cellcolor{gray!15}\textbf{0.745} & \cellcolor{gray!15}\textbf{0.755} & \cellcolor{gray!15}\textbf{0.667} & \cellcolor{gray!15}\textbf{0.793}\\

\multirow{2}{*}{Gemma2-9B-Instruct}
& Zero-shot & 0.681 & 0.635 & 0.649 & 0.639 & 0.702 & 0.681 & 0.689 & 0.516 & 0.762\\
& \cellcolor{gray!15}Dep-LLM & \cellcolor{gray!15}\textbf{0.787} & \cellcolor{gray!15}\textbf{0.748} & \cellcolor{gray!15}\textbf{0.766} & \cellcolor{gray!15}\textbf{0.755} & \cellcolor{gray!15}\textbf{0.798} & \cellcolor{gray!15}\textbf{0.787} & \cellcolor{gray!15}\textbf{0.791} & \cellcolor{gray!15}\textbf{0.667} & \cellcolor{gray!15}\textbf{0.844}\\

\multirow{2}{*}{Gemma3-12B-Base}
& Zero-shot & 0.702 & 0.653 & 0.665 & 0.657 & 0.715 & 0.702 & 0.707 & 0.533 & 0.781\\
& \cellcolor{gray!15}Dep-LLM & \cellcolor{gray!15}\textbf{0.787} & \cellcolor{gray!15}\textbf{0.763} & \cellcolor{gray!15}\textbf{0.807} & \cellcolor{gray!15}\textbf{0.770} & \cellcolor{gray!15}\textbf{0.829} & \cellcolor{gray!15}\textbf{0.787} & \cellcolor{gray!15}\textbf{0.795} & \cellcolor{gray!15}\textbf{0.706} & \cellcolor{gray!15}\textbf{0.833}\\

\multirow{2}{*}{Gemma3-12B-Instruct}
& Zero-shot & 0.702 & 0.690 & 0.726 & 0.685 & 0.767 & 0.702 & 0.715 & 0.611 & 0.759\\
& \cellcolor{gray!15}Dep-LLM & \cellcolor{gray!15}\textbf{0.851} & \cellcolor{gray!15}\textbf{0.826} & \cellcolor{gray!15}\textbf{0.812} & \cellcolor{gray!15}\textbf{0.818} & \cellcolor{gray!15}\textbf{0.849} & \cellcolor{gray!15}\textbf{0.851} & \cellcolor{gray!15}\textbf{0.849} & \cellcolor{gray!15}\textbf{0.741} & \cellcolor{gray!15}\textbf{0.896}\\

\bottomrule
\end{tabular}
\end{center}
\end{table*}

\section{Experiment}
\subsection{Experimental Settings}
\subsubsection{Datasets and Evaluation Metrics.} 

We evaluate Dep-LLM on two widely adopted clinical interview datasets for depression detection. The primary dataset is DAIC-WOZ~\cite{daic-gratch2014distress}, which contains semi-structured Wizard-of-Oz interviews collected for distress analysis. To further assess the generalizability of Dep-LLM beyond a single corpus, extended experiments are conducted on E-DAIC~\cite{Devault2014Simsensei}, an extended version of DAIC-WOZ in which the human operator behind the interviewer is totally replaced by an AI agent, leading to a noticeable distribution shift in interview style and participant behaviour.

To have a comprehensive and clinically meaningful insight on the performance, we report a panel of nine metrics covering  \textit{Accuracy} (Acc.), \textit{Macro Precision} (Ma*Prec.), \textit{Macro Recall} (Ma*Rec.), \textit{Macro F1} (Ma*F1.), \textit{Weighted-Average Precision} (WA*Prec.), \textit{Weighted-Average Recall} (WA*Rec.), \textit{Weighted-Average F1} (WA*F1.), \textit{Binary F1-Depressive} (F1-Depr.), \textit{Binary F1-Healthy} (F1-Heal.). Accuracy and Macro metrics reflect overall correctness while Weighted-Average metrics complementarily display class-balanced behaviour, and Per-class Binary metrics expose whether a method gains overall correctness at the cost of failing on the minority depressive class, following previous work~\cite{rinaldi2020predicting, Jung2024Hique, zhao-etal-2025-predicting}, due to the imbalanced labels on depressive and healthy of the two datasets.

\subsubsection{Implementation Details.}

We implement Dep-LLM in Pytorch on four NVIDIA H100 GPUs using the HuggingFace Transformers library. To verify that the performance gains brought by Dep-LLM are agnostic to the underlying foundation LLM, we evaluate it on a wide spectrum of open-source foundation LLMs that differ in family, scale and alignment status. Specifically, we cover Llama-2/3/4, Qwen-2.5/3/3.5 and Gemma-2/3 series, with parameter sizes ranging from $4$B to $17$B and variants covering base and instruction-tuned whenever available, yielding 21 foundation LLMs on DAIC-WOZ in total. For each foundation LLM, we compare Dep-LLM against vanilla zero-shot prompting baseline under identical generation hyperparameters (temperature, top-K, etc.). To ensure equality for zero-shot baselines, we designed a structured prompt for them as well as the Dep-LLM prompt, which are displayed in supplemental file.

\subsection{Comparison Experiment}
To demonstrate the effectiveness of Dep-LLM, we conducted three groups of comparison experiments: 1)~Dep-LLM versus vanilla zero-shot prompting on 21 foundation LLMs upon DAIC-WOZ; 2)~Dep-LLM versus vanilla zero-shot on 6 representative foundation LLMs upon E-DAIC to verify generalizability; 3)~Dep-LLM versus supervised domain-specific LLMs adapted to mental health field and the latest closed-source commercial LLMs upon DAIC-WOZ.

\subsubsection{Performance Against Zero-Shot Baselines on DAIC-WOZ}
Table~\ref{tab:Comparison} reports the head-to-head comparison between Dep-LLM and the zero-shot baseline on DAIC-WOZ across all 21 foundation LLMs. For readability, rows corresponding to Dep-LLM are shaded in \textit{gray}, and \textit{bold} marks the better result of the two methods for each foundation LLM. There are three discoveries to note: 1) Dep-LLM is universally superior on all metrics over zero-shot baseline. For example, Dep-LLM outperforms baseline on 19 out of 21 foundation LLMs on Ma*F1. and WA*F1., with only slight disadvantages for the two exceptions. Similar performance improvements are also observed on the remaining metrics, indicating the comprehensive superiority of Dep-LLM over zero-shot baselines. 2) regarding best performance, Dep-LLM paired with Gemma3-12B-Instruct attains the best scores on all metrics, reaching Ma*F1. of $0.818$, WA*F1. of $0.849$, F1-Depr. of $0.714$ and F1-Heal. of $0.896$. An average $12.6\%$ absolute improvement on the 9 metrics is remarkable for this foundation LLM. 3) such improvements are also significant on weaker models. For example, on Llama3-8B-Base, Dep-LLM has improved Ma*F1. by over $20\%$ absolute value, from $0.450$ to $0.666$, with other metrics improved greatly as well, suggesting that the structured analysis schema and confidence modulation of Dep-LLM bridge the gap of reasoning ability of weaker models as well, rather than merely amplifying already-strong ones. Overall, the table confirms that Dep-LLM is a generalizable and backbone-agnostic augmentation that systematically strengthens vanilla LLMs in long-context depression reasoning.

\begin{table*}[t]
\begin{center}
\renewcommand{\arraystretch}{1.00}
\caption{Extended Experiments on E-DAIC Dataset based on representative foundation models to demonstrate the generalizability of Dep-LLM.}
\label{tab:comp_exp}
\newcommand{\up}[1]{\;{\scriptsize\color{upcolor}{\textbf{$\uparrow$}#1}}} 
\begin{tabular}{ l | c | >{\centering\arraybackslash}p{1.06cm} | >{\centering\arraybackslash}p{1.06cm} | >{\centering\arraybackslash}p{1.06cm} | >{\centering\arraybackslash}p{1.06cm} | >{\centering\arraybackslash}p{1.06cm} | >{\centering\arraybackslash}p{1.06cm} | >{\centering\arraybackslash}p{1.06cm} | >{\centering\arraybackslash}p{1.06cm} | >{\centering\arraybackslash}p{1.06cm} }
\toprule
\textbf{Foundation LLM} & \textbf{Method} & \textbf{Acc.} & \textbf{Ma*Prec.} & \textbf{Ma*Rec.} & \textbf{Ma*F1.} & \textbf{WA*Prec.} & \textbf{WA*Rec.} & \textbf{WA*F1.} & \textbf{F1-Depr.} & \textbf{F1-Heal.}\\

\midrule
\multirow{2}{*}{Llama4-17B-Base}
& Zero-shot & 0.571 & 0.538 & 0.543 & 0.533 & 0.612 & 0.571 & 0.586 & 0.400 & 0.667\\
& \cellcolor{gray!15}Dep-LLM & \cellcolor{gray!15}\textbf{0.643} & \cellcolor{gray!15}\textbf{0.623} & \cellcolor{gray!15}\textbf{0.644} & \cellcolor{gray!15}\textbf{0.619} & \cellcolor{gray!15}\textbf{0.695} & \cellcolor{gray!15}\textbf{0.643} & \cellcolor{gray!15}\textbf{0.656} & \cellcolor{gray!15}\textbf{0.524} & \cellcolor{gray!15}\textbf{0.714}\\

\multirow{2}{*}{Llama4-17B-Instruct}
& Zero-shot & 0.571 & 0.524 & 0.526 & 0.522 & 0.599 & 0.617 & 0.631 & 0.471 & 0.700\\
& \cellcolor{gray!15}Dep-LLM & \cellcolor{gray!15}\textbf{0.661} & \cellcolor{gray!15}\textbf{0.604} & \cellcolor{gray!15}\textbf{0.607} & \cellcolor{gray!15}\textbf{0.605} & \cellcolor{gray!15}\textbf{0.666} & \cellcolor{gray!15}\textbf{0.661} & \cellcolor{gray!15}\textbf{0.663} & \cellcolor{gray!15}\textbf{0.457} & \cellcolor{gray!15}\textbf{0.753}\\

\midrule
\multirow{2}{*}{Qwen2.5-14B-Base}
& Zero-shot & \textbf{0.679} & 0.638 & 0.653 & 0.642 & 0.702 & \textbf{0.679} & 0.687 & 0.526 & \textbf{0.757}\\
& \cellcolor{gray!15}Dep-LLM & \cellcolor{gray!15}0.661 & \cellcolor{gray!15}\textbf{0.661} & \cellcolor{gray!15}\textbf{0.690} & \cellcolor{gray!15}\textbf{0.647} & \cellcolor{gray!15}\textbf{0.738} & \cellcolor{gray!15}0.661 & \cellcolor{gray!15}\textbf{0.674} & \cellcolor{gray!15}\textbf{0.578} & \cellcolor{gray!15}0.716\\

\multirow{2}{*}{Qwen2.5-14B-Instruct}
& Zero-shot & 0.696 & 0.662 & 0.683 & 0.666 & 0.725 & 0.696 & 0.705 & 0.564 & 0.767\\
& \cellcolor{gray!15}Dep-LLM & \cellcolor{gray!15}\textbf{0.732} & \cellcolor{gray!15}\textbf{0.708} & \cellcolor{gray!15}\textbf{0.741} & \cellcolor{gray!15}\textbf{0.711} & \cellcolor{gray!15}\textbf{0.774} & \cellcolor{gray!15}\textbf{0.732} & \cellcolor{gray!15}\textbf{0.742} & \cellcolor{gray!15}\textbf{0.634} & \cellcolor{gray!15}\textbf{0.789}\\

\midrule
\multirow{2}{*}{Gemma3-12B-Base}
& Zero-shot & 0.643 & 0.623 & 0.644 & 0.619 & 0.695 & 0.643 & 0.656 & 0.524 & 0.714\\
& \cellcolor{gray!15}Dep-LLM & \cellcolor{gray!15}\textbf{0.661} & \cellcolor{gray!15}\textbf{0.635} & \cellcolor{gray!15}\textbf{0.657} & \cellcolor{gray!15}\textbf{0.634} & \cellcolor{gray!15}\textbf{0.705} & \cellcolor{gray!15}\textbf{0.661} & \cellcolor{gray!15}\textbf{0.673} & \cellcolor{gray!15}\textbf{0.537} & \cellcolor{gray!15}\textbf{0.732}\\

\multirow{2}{*}{Gemma3-12B-Instruct}
& Zero-shot & 0.625 & 0.599 & 0.615 & 0.596 & 0.671 & 0.625 & 0.639 & 0.488 & 0.704\\
& \cellcolor{gray!15}Dep-LLM & \cellcolor{gray!15}\textbf{0.750} & \cellcolor{gray!15}\textbf{0.732} & \cellcolor{gray!15}\textbf{0.771} & \cellcolor{gray!15}\textbf{0.733} & \cellcolor{gray!15}\textbf{0.799} & \cellcolor{gray!15}\textbf{0.750} & \cellcolor{gray!15}\textbf{0.760} & \cellcolor{gray!15}\textbf{0.667} & \cellcolor{gray!15}\textbf{0.800}\\

\bottomrule
\end{tabular}
\end{center}
\end{table*}

\begin{table*}[t]
\begin{center}
\renewcommand{\arraystretch}{1.00}
\caption{Comparison Experiment against supervised domain-specific LLMs and closed-source commercial LLMs. For every metric, bold marks the best performance, and underline marks the second best performance.}
\label{tab:comp_external}
\newcommand{\up}[1]{\;{\scriptsize\color{upcolor}{\textbf{$\uparrow$}#1}}} 
\begin{tabular}{ l | >{\centering\arraybackslash}p{1.06cm} | >{\centering\arraybackslash}p{1.06cm} | >{\centering\arraybackslash}p{1.06cm} | >{\centering\arraybackslash}p{1.06cm} | >{\centering\arraybackslash}p{1.06cm} | >{\centering\arraybackslash}p{1.06cm} | >{\centering\arraybackslash}p{1.06cm} | >{\centering\arraybackslash}p{1.06cm} | >{\centering\arraybackslash}p{1.06cm} | >{\centering\arraybackslash}p{1.06cm} }
\toprule
\multirow{2}{*}{\textbf{Model}} 
& \multicolumn{5}{c|}{\textbf{DAIC-WOZ}} & \multicolumn{5}{c}{\textbf{E-DAIC}} \\
\cmidrule{2-11}
& \textbf{Acc.} & \textbf{Ma*F1.} & \textbf{WA*F1.} & \textbf{F1-Depr.} & \textbf{F1-Heal.} & \textbf{Acc.} & \textbf{Ma*F1.} & \textbf{WA*F1.} & \textbf{F1-Depr.} & \textbf{F1-Heal.}\\

\midrule
\multicolumn{11}{c}{Domain-specific LLMs.}\\

\midrule
MentalBERT & 0.766 & 0.726 & 0.768 & 0.621 & 0.831    & 0.714 & 0.701 & 0.726 & 0.636 & 0.765\\
MentalRoBERTa & 0.702 & 0.699 & 0.712 & 0.667 & 0.731 & \underline{0.732} & 0.711 & 0.742 & 0.634 & 0.789\\
ClinicalBERT & 0.702 & 0.677 & 0.714 & 0.588 & 0.767  & 0.696 & 0.666 & 0.705 & 0.564 & 0.767\\
MentaLLaMA & 0.660 & 0.631 & 0.673 & 0.529 & 0.733    & 0.696 & 0.646 & 0.699 & 0.514 & 0.779\\
MentalAlpaca & 0.681 & 0.650 & 0.692 & 0.545 & 0.754  & 0.678 & 0.663 & 0.691 & 0.591 & 0.735\\
BioMistral & 0.745 & 0.730 & 0.755 & 0.667 & 0.793    & 0.714 & 0.635 & 0.702 & 0.467 & \textbf{0.805}\\
Meditron & 0.745 & 0.695 & 0.745 & 0.571 & 0.818      & 0.714 & 0.689 & 0.724 & 0.600 & 0.778\\
\midrule

\multicolumn{11}{c}{Closed-Source Commercial LLMs.}\\

\midrule
GPT-5.5 & 0.702 & 0.691 & 0.715 & 0.632 & 0.750          & 0.696 & 0.679 & 0.708 & 0.605 & 0.754\\
Gemini-3.1-Pro & 0.702 & 0.668 & 0.711 & 0.563 & 0.774   & \underline{0.732} & \underline{0.725} & \underline{0.742} & \textbf{0.681} & 0.769\\
Claude-Opus-4.6 & \underline{0.787} & \underline{0.763} & \underline{0.794} & \underline{0.688} & \underline{0.839}  & 0.714 & 0.705 & 0.726 & 0.652 & 0.758\\
Grok-4.3 & 0.723 & 0.696 & 0.733 & 0.606 & 0.786         & 0.642 & 0.626 & 0.657 & 0.545 & 0.706\\
DeepSeek-V4 & 0.702 & 0.657 & 0.707 & 0.533 & 0.781      & 0.714 & 0.689 & 0.724 & 0.600 & 0.778\\

\midrule

\multicolumn{11}{c}{Our Method.}\\

\midrule

Dep-LLM* & \textbf{0.851} & \textbf{0.818} & \textbf{0.849} & \textbf{0.741} & \textbf{0.896} & \textbf{0.750} & \textbf{0.733} & \textbf{0.750} & \underline{0.667} & \underline{0.800}\\

\bottomrule

\end{tabular}
\end{center}
\end{table*}

\subsubsection{Performance Against Zero-Shot Baselines on E-DAIC}

To validate the generalizability of Dep-LLM across different corpora, we further evaluate representative foundation LLMs from Llama, Qwen and Gemma series on E-DAIC dataset. The results are reported in Table~\ref{tab:comp_exp}: 1) despite the difference of interview method and data quality, Dep-LLM continues to outperform the zero-shot baselines on all foundation models on Ma*Prec., Ma*Rec., Ma*F1., WA*Prec., WA*F1. and F1-Depr. and on 5 out of 6 models on Acc., WA*Rec., and F1-Heal. 2) The largest gain is observed on Gemma3-12B-Instruct where Dep-LLM raises Ma*F1. from $0.596$ to $0.733$, WA*F1. from $0.638$ to $0.760$, and F1-Depr. from $0.488$ to $0.667$. These results provide strong evidence for the effectiveness and generalizability of Dep-LLM across different foundation LLMs and different corpora.

\begin{table*}[t]
\begin{center}
\caption{Ablation Study to explore the importance of CoT reasoning, the included themes, and the multi-factor collaborative strategy. For every metric, bold marks the best performance, and underline marks the second best performance.}
\label{tab:ablation}
\newcommand{\up}[1]{\;{\scriptsize\color{upcolor}{\textbf{$\uparrow$}#1}}} 
\begin{tabular}{ l | >{\centering\arraybackslash}p{1.06cm} | >{\centering\arraybackslash}p{1.06cm} | >{\centering\arraybackslash}p{1.06cm} | >{\centering\arraybackslash}p{1.06cm} | >{\centering\arraybackslash}p{1.06cm} | >{\centering\arraybackslash}p{1.06cm} | >{\centering\arraybackslash}p{1.06cm} | >{\centering\arraybackslash}p{1.06cm} | >{\centering\arraybackslash}p{1.06cm} }

\toprule

\textbf{Model Variant} & \textbf{Acc.} & \textbf{Ma*Prec.} & \textbf{Ma*Rec.} & \textbf{Ma*F1.} & \textbf{WA*Prec.} & \textbf{WA*Rec.} & \textbf{WA*F1.} & \textbf{F1-Depr.} & \textbf{F1-Heal.}\\

\midrule

zero-shot & 0.702 & 0.690 & 0.726 & 0.685 & 0.767 & 0.702 & 0.715 & 0.611 & 0.759\\

\midrule

\multicolumn{10}{c}{Importance of Chain-of-Thought Reasoning.}\\

\midrule

w/o Multi-factor CoT & 0.702 & 0.644 & 0.644 & 0.644 & 0.702 & 0.702 & 0.702 & 0.500 & 0.788\\
w/o Thinking Steps & 0.702 & 0.664 & 0.685 & 0.668 & 0.730 & 0.702 & 0.711 & 0.563 & 0.774\\

\midrule

\multicolumn{10}{c}{Importance of Themes.}\\

\midrule

w/o Family & 0.787 & 0.748 & 0.766 & 0.755 & 0.798 & 0.787 & 0.791 & 0.667 & 0.844\\
w/o Work & 0.808 & 0.771 & 0.781 & 0.776 & 0.813 & 0.809 & 0.810 & 0.690 & 0.862\\
w/o Mental & 0.745 & 0.719 & 0.756 & 0.724 & 0.788 & 0.745 & 0.754 & 0.647 & 0.800\\
w/o Medical & 0.787 & 0.748 & 0.725 & 0.734 & 0.780 & 0.787 & 0.782 & 0.615 & 0.853\\
w/o Overall & \underline{0.830} & \underline{0.804} & 0.776 & \underline{0.787} & \underline{0.825} & \underline{0.830} & \underline{0.825} & 0.692 & \underline{0.882}\\

\midrule

\multicolumn{10}{c}{Multi-factor Collaborative Strategy.}\\

\midrule

Majority Voting & 0.702 & 0.676 & 0.706 & 0.678 & 0.747 & 0.702 & 0.714 & 0.588 & 0.767\\
High Possibility First & 0.702 & 0.676 & 0.706 & 0.678 & 0.747 & 0.702 & 0.714 & 0.588 & 0.767\\
High Confidence First & 0.745 & 0.709 & 0.736 & 0.716 & 0.771 & 0.745 & 0.752 & 0.625 & 0.806\\
Fusion w/o Confidence & 0.723 & 0.704 & 0.741 & 0.704 & 0.777 & 0.723 & 0.735 & 0.629 & 0.780\\
Fusion w/o Modulation & 0.808 & 0.773 & \underline{0.802} & 0.783 & \underline{0.825} & 0.809 & 0.813 & \underline{0.710} & 0.857\\

\midrule

\multicolumn{10}{c}{Full Version.}\\

\midrule

Dep-LLM* & \textbf{0.851} & \textbf{0.826} & \textbf{0.812} & \textbf{0.818} & \textbf{0.849} & \textbf{0.851} & \textbf{0.849} & \textbf{0.741} & \textbf{0.896}\\

\bottomrule

\end{tabular}
\end{center}
\end{table*}

\subsubsection{Performance Against other LLM methods} 

To evaluate Dep-LLM under broader landscape of LLM-based depression detection, we further compare it with two representative groups of strong competitors which is displayed in Table~\ref{tab:comp_external}: 1)~\textit{supervised domain-specific LLMs} that are explicitly pre-trained and fine-tuned on mental-health and clinical corpora, including MentalBERT~\cite{ji-etal-2022-mentalbert}, MentalRoBERTa~\cite{ji-etal-2022-mentalbert}, ClinicalBERT~\cite{huang2020clinicalbert}, MentaLLaMA~\cite{yang2023mentalllama}, MentalAlpaca~\cite{xu2024mental}, BioMistral~\cite{labrak2024biomistral} and Meditron~\cite{chen2023meditron}; 2)~\textit{closed-source commercial LLMs} including GPT-5.5, Gemini-3.1-Pro, Claude-Opus-4.6, Grok-4.3, and DeepSeek-V4, called under their public APIs. For a fair and reproducible comparison, Dep-LLM uses its best open-source foundation LLM Gemma3-12B-Instruct and is denoted as Dep-LLM* in the table. Table~\ref{tab:comp_external} reports five metrics (Acc., Ma*F1., WA*F1., F1-Depr., and F1-Heal.) for each method on DAIC-WOZ and E-DAIC dataset.

\paragraph{Versus Domain-Specific LLMs.} Although supervised domain-specific LLMs benefit from extensive mental health oriented pre-training or supervised fine-tuning, Dep-LLM nearly outperforms every one of them by clear difference on both datasets. On DAIC-WOZ, we compare the best performance of Dep-LLM* against them: Acc.($0.851$ vs.\ $0.766$), Ma*F1.($0.818$ vs.\ $0.730$), WA*F1.($0.849$ vs.\ $0.768$), F1-Depr.($0.741$ vs.\ $0.667$), F1-Heal.($0.896$ vs.\ $0.831$). On E-DAIC this advantage remains remarkable, except for tiny disadvantage ($0.5\%$) of the F1-Heal. on BioMistral, yet Dep-LLM* exceeds BioMistral on F1-Depr. by a big gap ($20.0\%$), which means BioMistral has an ill-balanced performance on E-DAIC. Crucially, Dep-LLM* achieves this without any task-specific training or fine-tuning, demonstrating that a well-structured training-free pipeline rivals and in fact surpasses supervised domain-specific models that depend on scarce labeled clinical data.

\paragraph{Versus Commercial LLMs.} The selected closed-source commercial LLMs represent the current frontier of general-purpose reasoning. On DAIC-WOZ, the strongest competitor is Claude-Opus-4.6 yet Dep-LLM* still wins on every metric with $+6.4\%$ Acc., $+5.5\%$ Ma*F1., $+5.5\%$ WA*F1., $+5.3\%$ F1-Depr., and $+5.7\%$ F1-Heal. On E-DAIC, Gemini-3.1-Pro emerges as the strongest commercial baseline. Yet Dep-LLM* outperforms Gemini-3.1-Pro on all metrics except for F1-Depr. with slight difference, indicating its stronger per-class specialization but not overall and balanced capability. It is worth noting that Dep-LLM* is built on a $12$B open-source foundation LLM, far smaller than the commercial systems in the comparison. Universally, the lightweight training-free framework beats the frontier commercial LLMs on most metrics and displays a more balanced performance, indicating that for high-stakes clinical scenarios, structured CoT decomposition and entropy-grounded confidence modulation are more critical than mere parameter scale.

\subsection{Ablation Study}\label{sec:ablation}
To demonstrate the effectiveness and indispensability of each component of Dep-LLM, we conduct a comprehensive ablation study on the foundation LLM Gemma3-12B-Instruct, reporting all nine metrics in Table~\ref{tab:ablation}. For readability, \textbf{bold} marks the best score for each metric (all achieved by full Dep-LLM*) and \underline{underline} marks the second best performance. The ablation study is organized into three groups, respectively manifesting the contribution of the Chain-of-Thought reasoning, each individual theme in the multi-factor schema, and the collaborative fusion strategy.

\subsubsection{Importance of Chain-of-Thought Reasoning}
We consider two decremental variants: \textit{w/o Multi-factor CoT}, which holistically evaluates the depressive/healthy/neutral possibilities and rationales from the raw clinical interview dialogue without structured multi-factor decomposition; and \textit{w/o Thinking Steps}, which removes the explicit step-by-step thinking prompt and directly generates multi-factor possibilities and rationales. The performance of both variants collapses to the zero-shot baseline with WA*F1. of $0.702$ and $0.711$ versus $0.849$ on full Dep-LLM*. F1-Depr. drops even more sharply from $0.741$ to $0.500$ and $0.563$, worse than zero-shot baseline. This result confirms that the CoT-driven structured multi-factor reasoning is a substantial foundation for the Dep-LLM framework due to its decomposition of the long and complex raw dialogues, without which Dep-LLM falls back to superficial holistic judgment.

\subsubsection{Importance of Themes}
We remove one single theme at a time alternatively and run prediction using the remaining four themes. It's evident from the results that every removal leads to a drop of performance. On average of five themes, Acc. drops by $5.96\%$, Ma*F1. drops by $6.28\%$ and WA*F1. drops by $5.66\%$. Among these themes, removal of mental state causes the biggest performance drop on most metrics, followed by medical history, family relationship and work satisfaction, which aligns with clinical intuition. Notably, while not so significant, removal of overall evaluation theme still hurts the performance, especially on Ma*Rec. and F1-Depr., meaning it's crucial for recognizing depressive samples, demonstrating that overall theme contributes supplementary integrative evidence from the four atomic themes rather than merely restating them. 

\subsubsection{Multi-factor Collaborative Strategy}
We compare our confidence-weighted fusion against three rule-based alternatives and two decremental variants. Rule-based strategies include \textit{Majority voting} adopting the label endorsed by the most themes, \textit{High Possibility First} adopting the label of the theme with highest possibility, and \textit{High Confidence First} adopting the label of the theme with highest confidence. They all underperform the full Dep-LLM* significantly, indicating that superficial rules without comprehensive utilization of possibilities and confidences are not reliable under complex clinical application. Two decremental variants include \textit{Fusion w/o Confidence} integrating themes only using possibilities, and \textit{Fusion w/o Modulation} integrating themes using raw confidence without intra-label and inter-theme modulation. Focusing on Acc. we can see that only possibility contributes $+2.1\%$ over zero-shot baseline, raw confidence contributes $+8.5\%$ over possibility, while modulation contributes the remaining $+4.3\%$ towards full Dep-LLM*. This improving trajectory is further explored in the following case study (Figure~\ref{fig:Case2}), quantitatively demonstrating that confidence mechanism and modulation adaptation are of great significance to the fusion process of the prediction.

\subsection{Case Study}\label{sec:case}
\begin{figure*}[ht]
\centering
\begin{center}
\centerline{\includegraphics[width=\textwidth]{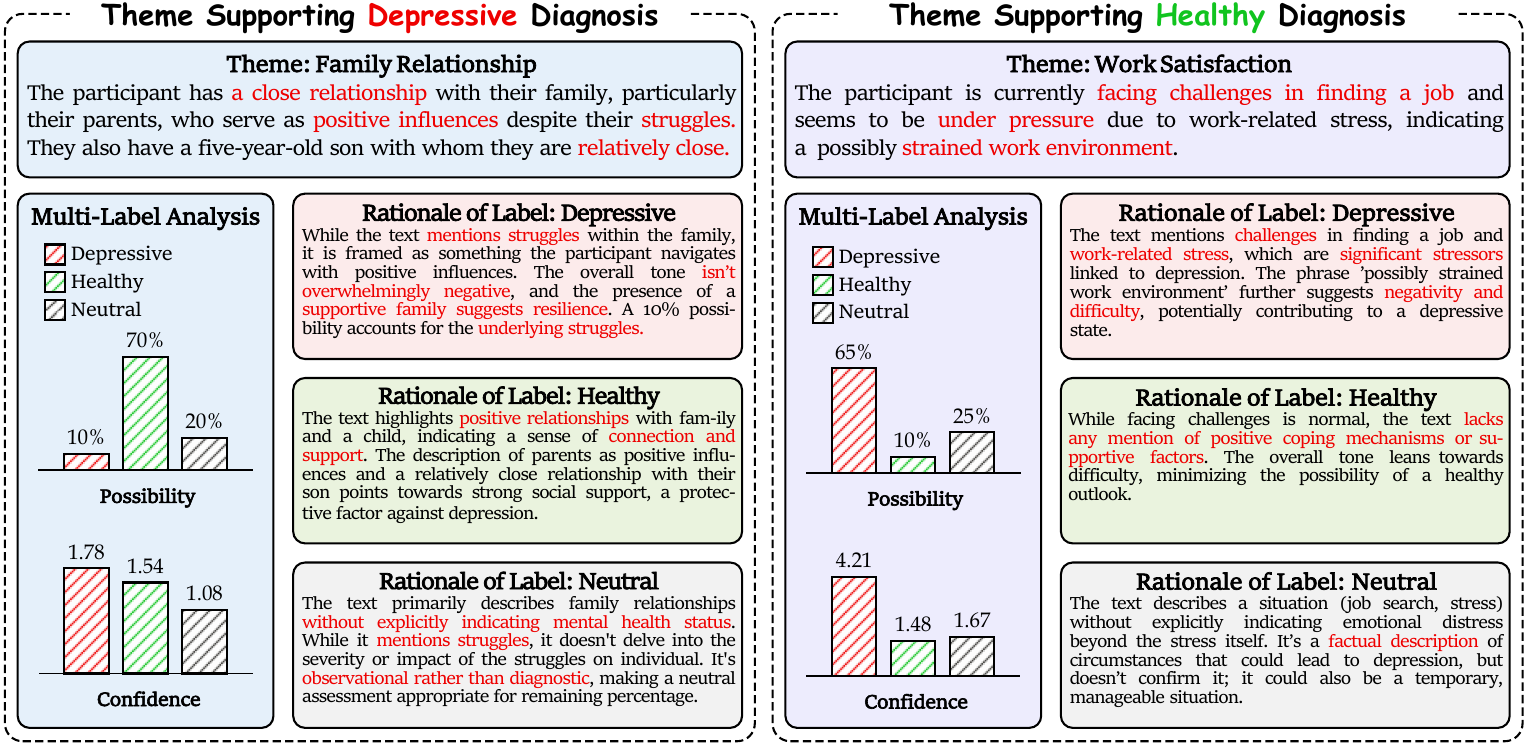}}
\caption{
Possibility, rationale and confidence in multi-factor Analysis of two case example.
}
\label{fig:Case1}
\end{center}
\end{figure*}

\subsubsection{Multi-factor Content Analysis}

To intuitively illustrate how Dep-LLM structurally decomposes clinical interviews and produces evidence-grounded reasoning, we visualize two representative thematic cases in Figure~\ref{fig:Case1}, respectively supporting a healthy and a depressive diagnosis. For each theme, Dep-LLM generates label-specific possibilities and rationales, during which confidences are calculated within the internal generation process, mirroring the analysis procedure of clinicians.

In the \textit{Family Relationship} case, the participant describes a close bond with their parents and son. Dep-LLM correctly identifies this as a protective signal and assigns highest possibility ($70\%$) to the healthy label. The depressive ($10\%$) and neutral ($20\%$) are assigned by rationales acknowledging the underlying struggles without overgeneralizing them, showing a clinically faithful attitude. In the \textit{Work Satisfaction} case, the participant reports troubles in finding a job and work-related stress. Dep-LLM assigns dominant possibility to the depressive label ($65\%$) and precisely captures clues like 'strained work environment', rather than superficial summaries. Meanwhile, the healthy rationale honestly admits the lack of positive evidence, and the neutral rationale appropriately frames the situation as 'a factual description' but not a confirming diagnostic factor. 

These two cases jointly demonstrate three merits of Dep-LLM. First, the structured multi-factor schema enables the framework to analyze based on clinically meaningful themes (family, work, etc.) rather than unclear holistic impressions. Second, the multi-label decomposition forces the LLM to argue for each diagnostic hypothesis, suppressing biased reasoning leaning towards one side. Third, the generated rationales are strictly evidence-anchored and can be traced back to specific clues in dialogue, which is essential for clinical reliability.

\begin{figure}[ht]
\centering
\begin{center}
\centerline{\includegraphics[width=\linewidth]{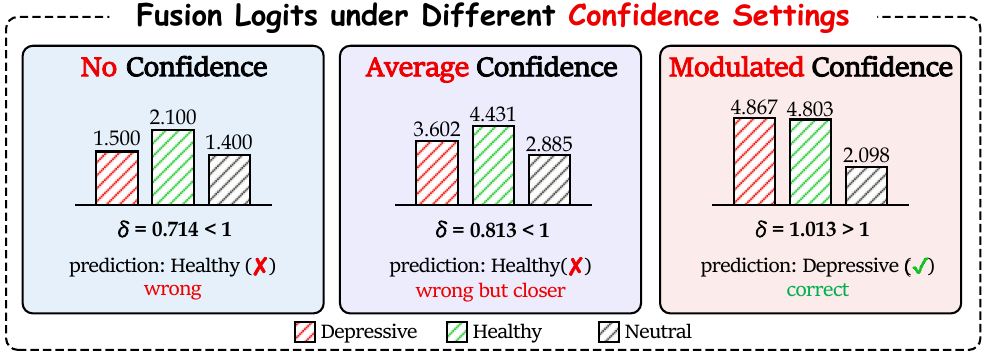}}
\caption{
Quantitative impact on prediction of confidence and modulation
}
\label{fig:Case2}
\end{center}
\end{figure}

\subsubsection{Confidence Mechanism}

To further display the role of the confidence mechanism in shaping the final diagnosis, in Figure~\ref{fig:Case2} we numerically visualize the fusion process under three different confidence settings. They are \textit{No Confidence} (all confidences degenerate to $1$), \textit{Average Confidence} (using raw average confidence $C^{avg}$, specified in Equation~\ref{eq:cavg}) and \textit{Modulated Confidence} (using $C^{mod}$ produced by intra-label and inter-theme modulation specified in Equation~\ref{eq:cmod}). These three settings are aligned with ablation study (Table~\ref{tab:ablation}), respectively corresponding to \textit{Fusion w/o Confidence}, \textit{Fusion w/o Modulation} and \textit{Full Version}, maintaining consistency.

Figure~\ref{fig:Case2} shows that fusion using only possibilities \textit{No Confidence} yields depression coefficient $\delta=0.714<1$ which mistakenly supports healthy diagnosis, because numerically large possibilities from less trustworthy themes have biased other informative depressive clues. \textit{Average Confidence} re-weights possibilities with $C^{avg}$ and produces $\delta=0.813<1$. Although the prediction is still incorrect, yet $\delta$ is nearer to the threshold $1$, indicating that average confidence carries a positive but insufficient signal when not normalized, where longer or more complex themes tend to dominate due to scale rather than reliability. Fusion under \textit{Modulated Confidence} exports $\delta=1.013>1$ and correctly predicts the depressive diagnosis, by allowing inter-theme interaction via amplifying a theme's weight when its own rationale is confident and other themes are uncertain.

The progress from incorrect to incorrect-but-closer to correct offers a clear quantitative trajectory of how confidence and modulation reshape the fusion process. The results verify the theory in Section~\ref{sec:conf} that raw average confidence derived from token entropy is a proxy for epistemic reliability, while contrastive modulation across labels and themes produces meaningful weights for fusion.

\subsection{Hyperparameter Sensitivity Analysis}\label{sec:hyp}

\begin{figure}[ht]
\centering
\begin{center}
\centerline{\includegraphics[width=\linewidth]{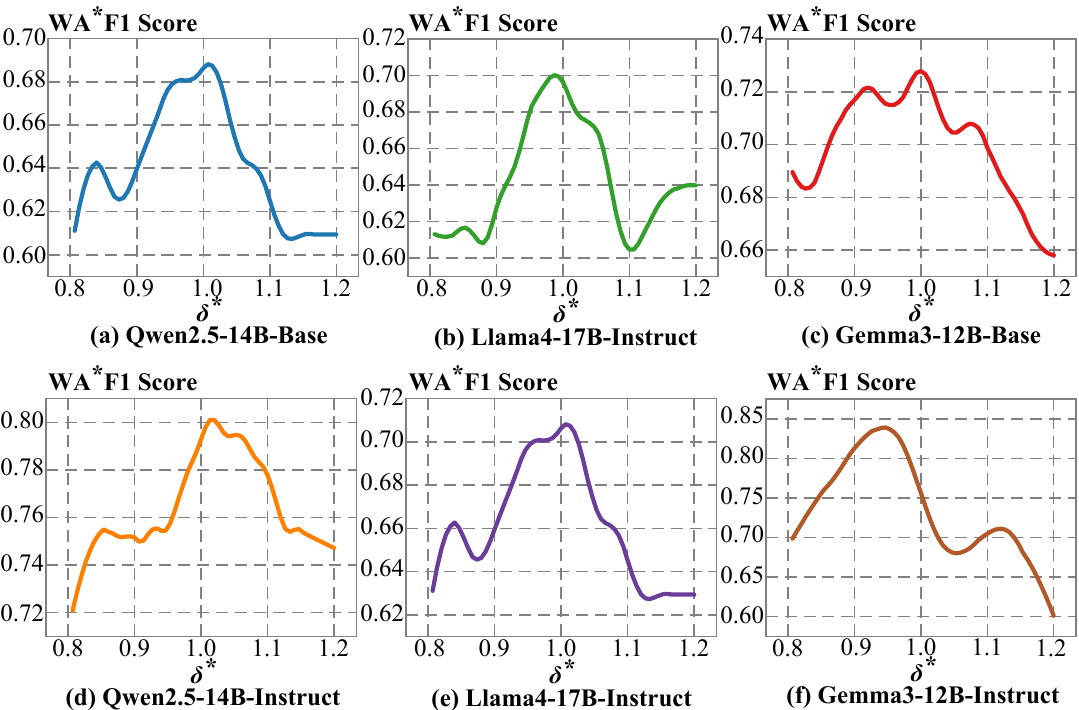}}
\caption{
The impact that value shift of $\delta^*$ exerts on WA*F1. Score of different models
}
\label{fig:DeltaShifts}
\end{center}
\end{figure}

\begin{figure}[ht]
\centering
\begin{center}
\centerline{\includegraphics[width=\linewidth]{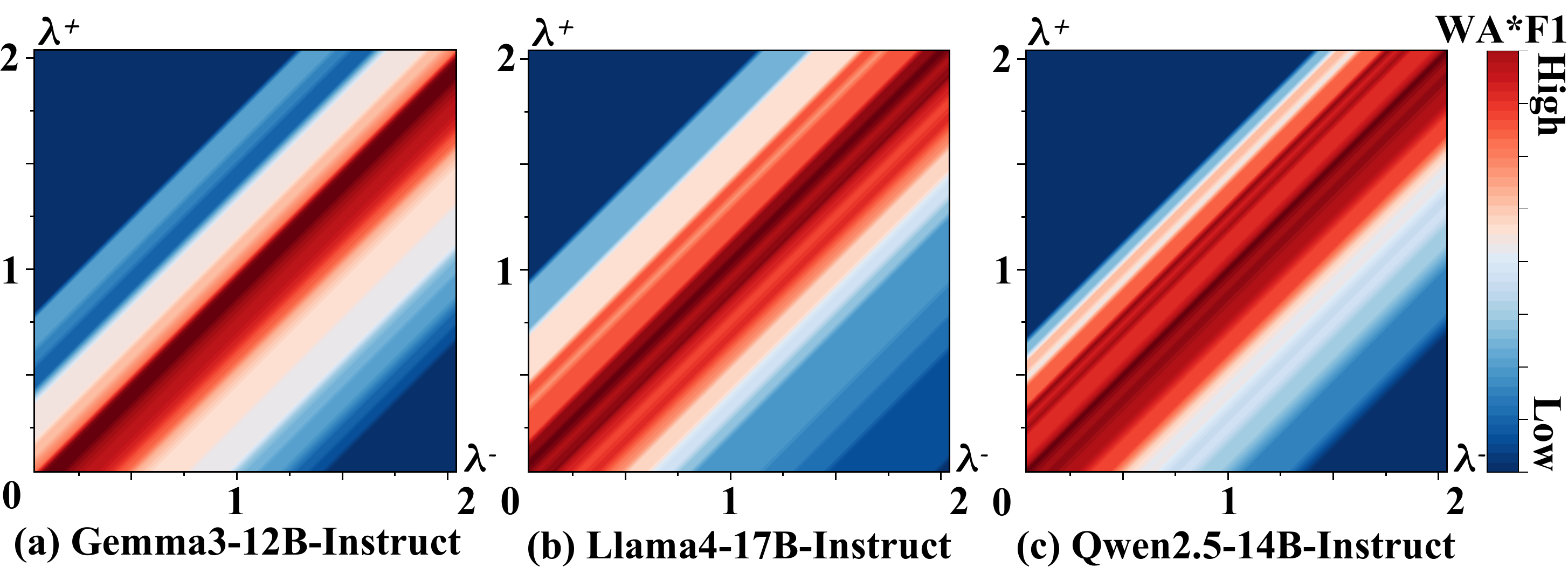}}
\caption{
The impact that value shift of $\lambda^+,\lambda^-$ exerts on WA*F1. Score of different models
}
\label{fig:LambdaShifts}
\end{center}
\end{figure}

Since Dep-LLM is a training-free framework, the two decision hyperparameters introduced in Section~\ref{sec:pred} --- depression threshold $\delta^*$ and neutral label calibrators $\lambda^+,\lambda^-$ --- are not learned from data but configured by design. To verify the robustness of the default configuration $\delta^*=1,\lambda^+=\lambda^-=1$ and to provide practical calibration guidance for real-world deployment, we conduct sensitivity analysis on these two hyperparameters across representative foundation LLMs from Qwen, Llama and Gemma series.

\subsubsection{Depression Threshold $\delta^*$.} We screen $\delta^*$ within the interval $[0.8,1.2]$ and report the resulting WA*F1. in Figure~\ref{fig:DeltaShifts}. We can draw three conclusions from the results. First, although the exact peak location varies across models, where Llama4-17B-Base and Gemma3-12B-Instruct perform best slightly below $1$ and the others peak slightly above $1$, yet they are all tightly clustered around $1$, confirming that $\delta^*=1$ is a near-optimal and well-justified default. Second, the WA*F1. curves remain relatively smooth around the optimum value, indicating that Dep-LLM is robust to small perturbations of $\delta^*$. Third, the discrepancy of the optimum value across models can be attributed to the intrinsic semantic preference of different LLMs, where the differences in architecture, pre-training corpora and alignment strategies introduce subtle but systematic biases towards specific diagnosis. Consequently, in deployment once the foundation LLM is fixed, $\delta^*$ may be further calibrated on a small held-out validation set by practitioners to gain additional accuracy by a marginal calibration effort.

\subsubsection{Neutral Label Calibrators $\lambda^+,\lambda^-$.} Analogously, we jointly screen $\lambda^+,\lambda^-$ within $[0,2]\times[0,2]$ and visualize the resulting WA*F1. as heatmaps in Figure~\ref{fig:LambdaShifts}. There we can see two patterns. First, the high-performance regions concentrate along the diagonal $\lambda^+=\lambda^-=1$ with slight shifts of different models, corresponding to our default symmetric setting in Section~\ref{sec:pred}. Second, there are many iso-performance bands with unit slope, which is predictable from the structure of Equation~\ref{eq:delta} that equal shifts of $\lambda^+$ and $\lambda^-$ would not change the value of $\delta$. In practice, $\lambda^+=\lambda^-=1$ remains a universal choice yet could be fine-calibrated for better performance.

Taken together, the sensitivity analysis on $\delta,\lambda^+,\lambda^-$ demonstrates that Dep-LLM is robust by design with default hyperparameter configuration grounded in both empirical and experimental analysis. Furthermore, Dep-LLM still allows clinicians to calibrate them for deployment-specific adaptation in a transparent and lightweight approach, which is usually critical in high-stakes mental health applications.

\section{Conclusion}
In this paper, we present \textbf{Dep-LLM}, a training-free framework for automatic depression detection that jointly addresses two long-standing challenges: inaccurate reasoning over sparse clues in long dialogues, along with the data scarcity and high training cost of supervised systems. Mirroring the diagnostic workflow of clinicians, Dep-LLM operates entirely on frozen LLMs through coordinated stages: 1) a CoT Depression Multi-factor Analysis module that decomposes each interview into SCID-aligned themes and extracts evidence-grounded rationales; 2) a Confidence Analysis and Modulation module that estimates the epistemic reliability of each rationale from token-level entropy and modulates it across labels and themes; 3) a Collaborative Multi-factor Fusion that integrates these confidence-weighted signals into a transparent diagnosis. Since the pipeline introduces no learnable parameters and requires no labeled data, it can be deployed under the data and computational constraints typical of real clinical settings. Extensive experiments on DAIC-WOZ and E-DAIC show that Dep-LLM consistently improves diverse foundation LLMs over zero-shot baseline and outperforms both supervised domain-specific LLMs and the latest commercial LLMs, while retaining high interpretability.

\bibliographystyle{IEEEtran}
\bibliography{TAFFC_custom.bib}

@book{apa2013diagnostic,
  title={Diagnostic and Statistical Manual of Mental Disorders (5th ed.)},
  author={{American Psychiatric Association}},
  publisher={American Psychiatric Publishing},
  address={Arlington, VA},
  year={2013},
  url={https://doi.org/10.1176/appi.books.9780890425596}
}

@manual{who2024world,
  title={World Mental Health Today and Mental Health Atlas 2024},
  author={{World Health Organization}},
  organization={World Health Organization},
  year={2024},
  address={Geneva},
  url={https://www.who.int/publications/i/item/9789240114487}
}

@inproceedings{loweimi2025zero,
  title={Zero-Shot Speech-Based Depression and Anxiety Assessment with LLMs},
  author={Loweimi, Erfan and Others},
  booktitle={Proceedings of Interspeech 2025},
  year={2025},
  pages={489--493},
  url={https://www.research.ed.ac.uk/en/publications/zero-shot-speech-based-depression-and-anxiety-assessment-with-llm}
}

@article{teferra2024leveraging,
  title={Leveraging Large Language Models for Automated Depression Screening},
  author={Teferra, Bazen Gashaw and Perivolaris, Argyrios},
  journal={Frontiers in Psychiatry},
  year={2024},
  volume={15},
  url={https://pmc.ncbi.nlm.nih.gov/articles/PMC12303271/}
}

@inproceedings{ge2025survey,
   title={A Survey of Large Language Models in Mental Health Disorder Detection on Social Media},
   url={http://dx.doi.org/10.1109/ICDEW67478.2025.00027},
   DOI={10.1109/icdew67478.2025.00027},
   booktitle={2025 IEEE 41st International Conference on Data Engineering Workshops (ICDEW)},
   publisher={IEEE},
   author={Ge, Zhuohan and Hu, Nicole and Li, Darian and Wang, Yubo and Qi, Shihao and Xu, Yuming and Shi, Han and Zhang, Jason},
   year={2025},
   month=may, pages={164–176} }

@article{mao2025enhancing,
  title={Enhancing TextGCN for Depression Detection on Social Media with Emotion Representation},
  author={Mao, H and Han, Q},
  journal={Frontiers in Psychology},
  volume={16},
  pages={1612769},
  year={2025},
  url={https://www.frontiersin.org/journals/psychology/articles/10.3389/fpsyg.2025.1612769/full}
}

@article{zogan2022explainable,
  title={Explainable Depression Detection with Multi-Aspect Features Using a Hybrid Deep Learning Model on Social Media},
  author={Zogan, Hamad and Razzak, Imran and Wang, Xianzhi and Jameel, Shoaib and Xu, Guandong},
  journal={World Wide Web},
  volume={25},
  number={1},
  pages={281--304},
  year={2022},
  url={https://link.springer.com/article/10.1007/s11280-021-00992-2}
}

@InProceedings{Yao2024Depression,
author = {Yao, Xingda and Ying, Lingkang and He, Tianxiang and Ren, Ligang and Xu, Ruiji and Mao, Keji},
title = {Depression Detection Based on Multilevel Semantic Features},
year = {2024},
isbn = {978-3-031-72352-0},
publisher = {Springer-Verlag},
address = {Berlin, Heidelberg},
url = {https://doi.org/10.1007/978-3-031-72353-7_4},
doi = {10.1007/978-3-031-72353-7_4},
booktitle = {Artificial Neural Networks and Machine Learning – ICANN 2024: 33rd International Conference on Artificial Neural Networks, Lugano, Switzerland, September 17–20, 2024, Proceedings, Part VIII},
pages = {44–55},
numpages = {12},
location = {Lugano, Switzerland}
}

@article {kim2025medical,
	author = {Kim, Yubin and Jeong, Hyewon and Chen, Shan and Li, Shuyue Stella and Lu, Mingyu and Alhamoud, Kumail and Mun, Jimin and Grau, Cristina and Jung, Minseok and Gameiro, Rodrigo and Fan, Lizhou and Park, Eugene and Lin, Tristan and Yoon, Joonsik and Yoon, Wonjin and Sap, Maarten and Tsvetkov, Yulia and Liang, Paul and Xu, Xuhai and Liu, Xin and McDuff, Daniel and Lee, Hyeonhoon and Park, Hae Won and Tulebaev, Samir and Breazeal, Cynthia},
	title = {Medical Hallucination in Foundation Models and Their Impact on Healthcare},
	elocation-id = {2025.02.28.25323115},
	year = {2025},
	doi = {10.1101/2025.02.28.25323115},
	publisher = {Cold Spring Harbor Laboratory Press},
	URL = {https://www.medrxiv.org/content/early/2025/03/03/2025.02.28.25323115},
	eprint = {https://www.medrxiv.org/content/early/2025/03/03/2025.02.28.25323115.full.pdf},
	journal = {medRxiv}
}

@article{Zhou2025large,
  author = {Zhou, Juexiao and Li, Haoyang and Chen, Siyuan and Chen, Zhangtianyi and Han, Zhongyi and Gao, Xin},
  title = {Large language models in biomedicine and healthcare},
  journal = {npj Artificial Intelligence},
  year = {2025},
  month = {dec},
  volume = {1},
  number = {1},
  pages = {44},
  doi = {10.1038/s44387-025-00047-1},
  url = {https://doi.org/10.1038/s44387-025-00047-1},
  issn = {3005-1460},
}

@Article{miao2024integrating,
AUTHOR = {Miao, Jing and Thongprayoon, Charat and Suppadungsuk, Supawadee and Garcia Valencia, Oscar A. and Cheungpasitporn, Wisit},
TITLE = {Integrating Retrieval-Augmented Generation with Large Language Models in Nephrology: Advancing Practical Applications},
JOURNAL = {Medicina},
VOLUME = {60},
YEAR = {2024},
NUMBER = {3},
ARTICLE-NUMBER = {445},
URL = {https://www.mdpi.com/1648-9144/60/3/445},
PubMedID = {38541171},
ISSN = {1648-9144},
DOI = {10.3390/medicina60030445}
}

@article{xu2024mental,
  title={Mental-LLM: Leveraging Large Language Models for Mental Health Prediction via Online Text Data},
  author={Xu, Xuhai and Yao, Bingsheng and Dong, Yuanzhe and Gabriel, Saadia and Yu, Hong and Hendler, James and Ghassemi, Marzyeh and Dey, Anind K and Wang, Dakuo},
  journal={Proceedings of the ACM on Interactive, Mobile, Wearable and Ubiquitous Technologies},
  volume={8},
  number={1},
  pages={1--32},
  year={2024},
  url={https://arxiv.org/abs/2307.14385}
}

@inproceedings{yang2023mentalllama,
   title={MentaLLaMA: Interpretable Mental Health Analysis on Social Media with Large Language Models},
   url={http://dx.doi.org/10.1145/3589334.3648137},
   DOI={10.1145/3589334.3648137},
   booktitle={Proceedings of the ACM Web Conference 2024},
   publisher={ACM},
   author={Yang, Kailai and Zhang, Tianlin and Kuang, Ziyan and Xie, Qianqian and Huang, Jimin and Ananiadou, Sophia},
   year={2024},
   month=may, pages={4489–4500},
   collection={WWW ’24}
}

@inproceedings{wei2022chain,
  title={Chain-of-Thought Prompting Elicits Reasoning in Large Language Models},
  author={Wei, Jason and Wang, Xuezhi and Schuurmans, Dale and Bosma, Maarten and Chi, Ed and Le, Quoc and Zhou, Denny},
  booktitle={Advances in Neural Information Processing Systems},
  volume={35},
  pages={24824--24837},
  year={2022},
  url={https://arxiv.org/abs/2201.11903}
}

@article{teng2025enhancing,
  title={Enhancing Depression Detection with Chain-of-Thought Prompting: From Emotion to Reasoning Using Large Language Models},
  author={Teng, Shiyu and Liu, Jiaqing and Jain, Rahul Kumar and Chai, Shurong and Hou, Ruibo and Tateyama, Tomoko and Lin, Lanfen and Chen, Yen-Wei},
  journal={arXiv preprint arXiv:2502.05879},
  year={2025},
  url={https://arxiv.org/abs/2502.05879}
}

@misc{wu2024uncertainty,
      title={Uncertainty Estimation of Large Language Models in Medical Question Answering}, 
      author={Jiaxin Wu and Yizhou Yu and Hong-Yu Zhou},
      year={2024},
      eprint={2407.08662},
      archivePrefix={arXiv},
      primaryClass={cs.CL},
      url={https://arxiv.org/abs/2407.08662}, 
}

@article{hu2025agentmental,
  title={AgentMental: An Interactive Multi-Agent Framework for Explainable and Adaptive Mental Health Assessment},
  author={Hu, Jinpeng and Others},
  journal={arXiv preprint arXiv:2508.11567},
  year={2025},
  url={https://arxiv.org/abs/2508.11567}
}

@inproceedings{ji-etal-2022-mentalbert,
    title = "{M}ental{BERT}: Publicly Available Pretrained Language Models for Mental Healthcare",
    author = "Ji, Shaoxiong  and
      Zhang, Tianlin  and
      Ansari, Luna  and
      Fu, Jie  and
      Tiwari, Prayag  and
      Cambria, Erik",
    editor = "Calzolari, Nicoletta  and
      B{\'e}chet, Fr{\'e}d{\'e}ric  and
      Blache, Philippe  and
      Choukri, Khalid  and
      Cieri, Christopher  and
      Declerck, Thierry  and
      Goggi, Sara  and
      Isahara, Hitoshi  and
      Maegaard, Bente  and
      Mariani, Joseph  and
      Mazo, H{\'e}l{\`e}ne  and
      Odijk, Jan  and
      Piperidis, Stelios",
    booktitle = "Proceedings of the Thirteenth Language Resources and Evaluation Conference",
    month = jun,
    year = "2022",
    address = "Marseille, France",
    publisher = "European Language Resources Association",
    url = "https://aclanthology.org/2022.lrec-1.778/",
    pages = "7184--7190",
}

@article{brodey2018rapid,
  title={Rapid and Accurate Diagnosis of Mental Disorders in the General Population: Validity of the Computerized Adaptive Testing–Mental Health (CAT-MH) Module},
  author={Brodey, Benjamin B and Purcell, Susan E and Rhea, Karen and Maier, Philip and First, Michael B and Zweede, Lisa and Sinisterra, Manuela and Nunn, M Brad and Austin, Marie-Paule and Brodey, Inger S},
  journal={Journal of Medical Internet Research},
  year={2018},
  volume={20},
  number={3},
  pages={e10685},
  url={https://www.ncbi.nlm.nih.gov/pmc/articles/PMC5889494/}
}

@article{shankman2018reliability,
  title={Reliability and Validity of Severity Dimensions of Psychopathology Assessed Using the Structured Clinical Interview for DSM-5 (SCID)},
  author={Shankman, Stewart A and Funkhouser, Carter J and Klein, Daniel N and Davila, Joanne and Lerner, Debra and Hee, Danelle},
  journal={International Journal of Methods in Psychiatric Research},
  year={2018},
  volume={27},
  number={1},
  pages={e1590},
  url={https://pubmed.ncbi.nlm.nih.gov/29034525/}
}

@inproceedings{rinaldi2020predicting,
    title = "Predicting Depression in Screening Interviews from Latent Categorization of Interview Prompts",
    author = "Rinaldi, Alex  and
      Fox Tree, Jean  and
      Chaturvedi, Snigdha",
    editor = "Jurafsky, Dan  and
      Chai, Joyce  and
      Schluter, Natalie  and
      Tetreault, Joel",
    booktitle = "Proceedings of the 58th Annual Meeting of the Association for Computational Linguistics",
    month = jul,
    year = "2020",
    address = "Online",
    publisher = "Association for Computational Linguistics",
    url = "https://aclanthology.org/2020.acl-main.2/",
    doi = "10.18653/v1/2020.acl-main.2",
    pages = "7--18"
}

@inproceedings{zhao-etal-2025-predicting,
    title = "Predicting Depression in Screening Interviews from Interactive Multi-Theme Collaboration",
    author = "Zhao, Xianbing  and
      Lyu, Yiqing  and
      Wang, Di  and
      Tang, Buzhou",
    editor = "Che, Wanxiang  and
      Nabende, Joyce  and
      Shutova, Ekaterina  and
      Pilehvar, Mohammad Taher",
    booktitle = "Findings of the Association for Computational Linguistics: ACL 2025",
    month = jul,
    year = "2025",
    address = "Vienna, Austria",
    publisher = "Association for Computational Linguistics",
    url = "https://aclanthology.org/2025.findings-acl.1181/",
    doi = "10.18653/v1/2025.findings-acl.1181",
    pages = "23025--23035",
    ISBN = "979-8-89176-256-5"
}

@inproceedings{daic-gratch2014distress,
  title={The distress analysis interview corpus of human and computer interviews.},
  author={Gratch, Jonathan and Artstein, Ron and Lucas, Gale M and Stratou, Giota and Scherer, Stefan and Nazarian, Angela and Wood, Rachel and Boberg, Jill and DeVault, David and Marsella, Stacy and others},
  booktitle={LREC},
  pages={3123--3128},
  year={2014},
  organization={Reykjavik},
  url={https://api.semanticscholar.org/CorpusID:14488823}
}

@inproceedings{Devault2014Simsensei,
author = {DeVault, David and Artstein, Ron and Benn, Grace and Dey, Teresa and Fast, Ed and Gainer, Alesia and Georgila, Kallirroi and Gratch, Jon and Hartholt, Arno and Lhommet, Margaux and Lucas, Gale and Marsella, Stacy and Morbini, Fabrizio and Nazarian, Angela and Scherer, Stefan and Stratou, Giota and Suri, Apar and Traum, David and Wood, Rachel and Xu, Yuyu and Rizzo, Albert and Morency, Louis-Philippe},
title = {SimSensei kiosk: a virtual human interviewer for healthcare decision support},
year = {2014},
isbn = {9781450327381},
publisher = {International Foundation for Autonomous Agents and Multiagent Systems},
address = {Richland, SC},
booktitle = {Proceedings of the 2014 International Conference on Autonomous Agents and Multi-Agent Systems},
pages = {1061–1068},
numpages = {8},
location = {Paris, France},
series = {AAMAS '14},
url= {https://dl.acm.org/doi/10.5555/2615731.2617415}
}

@inproceedings{Jung2024Hique,
author = {Jung, Juho and Kang, Chaewon and Yoon, Jeewoo and Kim, Seungbae and Han, Jinyoung},
title = {HiQuE: Hierarchical Question Embedding Network for Multimodal Depression Detection},
year = {2024},
isbn = {9798400704369},
publisher = {Association for Computing Machinery},
address = {New York, NY, USA},
url = {https://doi.org/10.1145/3627673.3679797},
doi = {10.1145/3627673.3679797},
booktitle = {Proceedings of the 33rd ACM International Conference on Information and Knowledge Management},
pages = {1049–1059},
numpages = {11},
location = {Boise, ID, USA},
series = {CIKM '24}
}

@inproceedings{zadeh2017tensor,
  title = "Tensor Fusion Network for Multimodal Sentiment Analysis",
    author = "Zadeh, Amir  and
      Chen, Minghai  and
      Poria, Soujanya  and
      Cambria, Erik  and
      Morency, Louis-Philippe",
    editor = "Palmer, Martha  and
      Hwa, Rebecca  and
      Riedel, Sebastian",
    booktitle = "Proceedings of the 2017 Conference on Empirical Methods in Natural Language Processing",
    month = sep,
    year = "2017",
    address = "Copenhagen, Denmark",
    publisher = "Association for Computational Linguistics",
    url = "https://aclanthology.org/D17-1115/",
    doi = "10.18653/v1/D17-1115",
    pages = "1103--1114"
}

@inproceedings{gong2017topic,
  author = {Gong, Yuan and Poellabauer, Christian},
  title = {Topic Modeling Based Multi-modal Depression Detection},
  year = {2017},
  isbn = {9781450355025},
  publisher = {Association for Computing Machinery},
  address = {New York, NY, USA},
  url = {https://doi.org/10.1145/3133944.3133945},
  doi = {10.1145/3133944.3133945},
  booktitle = {Proceedings of the 7th Annual Workshop on Audio/Visual Emotion Challenge},
  pages = {69–76},
  numpages = {8},
  location = {Mountain View, California, USA},
  series = {AVEC '17}
}

@inproceedings{mallol2019hierarchical,
  author = {Mallol-Ragolta, Adria and Zhao, Ziping and Stappen, Lukas and Schuller, Björn},
  year = {2019},
  month = {09},
  pages = {221-225},
  title = {A Hierarchical Attention Network-Based Approach for Depression Detection from Transcribed Clinical Interviews},
  doi = {10.21437/Interspeech.2019-2036}
}

@inproceedings{tsai2019multimodal,
  title = "Multimodal Transformer for Unaligned Multimodal Language Sequences",
    author = "Tsai, Yao-Hung Hubert  and
      Bai, Shaojie  and
      Liang, Paul Pu  and
      Kolter, J. Zico  and
      Morency, Louis-Philippe  and
      Salakhutdinov, Ruslan",
    editor = "Korhonen, Anna  and
      Traum, David  and
      M{\`a}rquez, Llu{\'i}s",
    booktitle = "Proceedings of the 57th Annual Meeting of the Association for Computational Linguistics",
    month = jul,
    year = "2019",
    address = "Florence, Italy",
    publisher = "Association for Computational Linguistics",
    url = "https://aclanthology.org/P19-1656/",
    doi = "10.18653/v1/P19-1656",
    pages = "6558--6569"
}

@misc{hazarika2020misa,
  title={MISA: Modality-Invariant and -Specific Representations for Multimodal Sentiment Analysis}, 
      author={Devamanyu Hazarika and Roger Zimmermann and Soujanya Poria},
      year={2020},
      eprint={2005.03545},
      archivePrefix={arXiv},
      primaryClass={cs.CL},
      url={https://arxiv.org/abs/2005.03545}
}

@inproceedings{ye2024depmamba,
  author={Ye, Jiaxin and Zhang, Junping and Shan, Hongming},
  booktitle={ICASSP 2025 - 2025 IEEE International Conference on Acoustics, Speech and Signal Processing (ICASSP)}, 
  title={DepMamba: Progressive Fusion Mamba for Multimodal Depression Detection}, 
  year={2025},
  volume={},
  number={},
  pages={1-5},
  doi={10.1109/ICASSP49660.2025.10889975}
}

@article{yang2024mmpf,
  author={Yang, Biao and Cao, Miaomiao and Zhu, Xianlin and Wang, Suhong and Yang, Changchun and Ni, Rongrong and Liu, Xiaofeng},
  journal={IEEE Transactions on Computational Social Systems}, 
  title={MMPF: Multimodal Purification Fusion for Automatic Depression Detection}, 
  year={2024},
  volume={11},
  number={6},
  pages={7421-7434},
  doi={10.1109/TCSS.2024.3411616}
}

@article{chen2025ttfnet,
  author={Chen, Xiyuan and Shao, Zhuhong and Jiang, Yinan and Chen, Runsen and Wang, Yunlong and Li, Bicao and Niu, Mingyue and Chen, Hongguang and Hu, Qiang and Wu, Jiasong and Yang, Chunfeng and Shang, Yuanyuan},
  journal={IEEE Journal of Biomedical and Health Informatics}, 
  title={TTFNet: Temporal-Frequency Features Fusion Network for Speech Based Automatic Depression Recognition and Assessment}, 
  year={2025},
  volume={29},
  number={10},
  pages={7536-7548},
  doi={10.1109/JBHI.2025.3574864}
}

@article{flores2025wavface,
  author={Flores, Ricardo and Tlachac, ML and Shrestha, Avantika and Rundensteiner, Elke A.},
  journal={IEEE Journal of Biomedical and Health Informatics}, 
  title={WavFace: A Multimodal Transformer-Based Model for Depression Screening}, 
  year={2025},
  volume={29},
  number={5},
  pages={3632-3641},
  doi={10.1109/JBHI.2025.3529348}
}

@article{tao2024depmstat,
  author={Tao, Yongfeng and Yang, Minqiang and Li, Huiru and Wu, Yushan and Hu, Bin},
  journal={IEEE Transactions on Knowledge and Data Engineering}, 
  title={DepMSTAT: Multimodal Spatio-Temporal Attentional Transformer for Depression Detection}, 
  year={2024},
  volume={36},
  number={7},
  pages={2956-2966},
  doi={10.1109/TKDE.2024.3350071}
}

@inproceedings{chen2024depression,
  title = "Depression Detection in Clinical Interviews with {LLM}-Empowered Structural Element Graph",
  author = "Chen, Zhuang  and
      Deng, Jiawen  and
      Zhou, Jinfeng  and
      Wu, Jincenzi  and
      Qian, Tieyun  and
      Huang, Minlie",
  editor = "Duh, Kevin  and
      Gomez, Helena  and
      Bethard, Steven",
  booktitle = "Proceedings of the 2024 Conference of the North American Chapter of the Association for Computational Linguistics: Human Language Technologies (Volume 1: Long Papers)",
  month = jun,
  year = "2024",
  address = "Mexico City, Mexico",
  publisher = "Association for Computational Linguistics",
  url = "https://aclanthology.org/2024.naacl-long.452/",
  doi = "10.18653/v1/2024.naacl-long.452",
  pages = "8181--8194"
}

@article{ohse2024zeroshot,
  title = {Zero-Shot Strike: Testing the generalisation capabilities of out-of-the-box LLM models for depression detection},
  journal = {Computer Speech \& Language},
  volume = {88},
  pages = {101663},
  year = {2024},
  issn = {0885-2308},
  doi = {https://doi.org/10.1016/j.csl.2024.101663},
  url = {https://www.sciencedirect.com/science/article/pii/S0885230824000469},
  author = {Julia Ohse and Bakir Hadžić and Parvez Mohammed and Nicolina Peperkorn and Michael Danner and Akihiro Yorita and Naoyuki Kubota and Matthias Rätsch and Youssef Shiban}
}

@misc{zheng2025towards,
  title={Towards Explainable Multimodal Depression Recognition for Clinical Interviews}, 
  author={Wenjie Zheng and Qiming Xie and Zengzhi Wang and Jianfei Yu and Rui Xia},
  year={2025},
  eprint={2501.16106},
  archivePrefix={arXiv},
  primaryClass={cs.CL},
  url={https://arxiv.org/abs/2501.16106}
}

@inproceedings{lan2025depression,
  title = "Depression Detection on Social Media with Large Language Models",
  author = "Lan, Xiaochong  and
      Han, Zhiguang  and
      Cheng, Yiming  and
      Sheng, Li  and
      Feng, Jie  and
      Gao, Chen  and
      Li, Yong",
  editor = "Potdar, Saloni  and
      Rojas-Barahona, Lina  and
      Montella, Sebastien",
  booktitle = "Proceedings of the 2025 Conference on Empirical Methods in Natural Language Processing: Industry Track",
  month = nov,
  year = "2025",
  address = "Suzhou (China)",
  publisher = "Association for Computational Linguistics",
  url = "https://aclanthology.org/2025.emnlp-industry.151/",
  doi = "10.18653/v1/2025.emnlp-industry.151",
  pages = "2155--2171",
  ISBN = "979-8-89176-333-3"
}

@inproceedings{zhou2024intermind,
  author = {Zhou, Zhiyuan and Liu, Jilong and Wang, Sanwang and Hao, Shijie and Guo, Yanrong and Hong, Richang},
  title = {InterMind: Doctor-Patient-Family Interactive Depression Assessment Empowered by Large Language Models},
  year = {2025},
  isbn = {9798400720352},
  publisher = {Association for Computing Machinery},
  address = {New York, NY, USA},
  url = {https://doi.org/10.1145/3746027.3754755},
  doi = {10.1145/3746027.3754755},
  booktitle = {Proceedings of the 33rd ACM International Conference on Multimedia},
  pages = {5480–5489},
  numpages = {10},
  location = {Dublin, Ireland},
  series = {MM '25}
}

@inproceedings{liu2025eeyore,
  title = "Eeyore: Realistic Depression Simulation via Expert-in-the-Loop Supervised and Preference Optimization",
  author = "Liu, Siyang  and
      Brie, Bianca  and
      Li, Wenda  and
      Biester, Laura  and
      Lee, Andrew  and
      Pennebaker, James  and
      Mihalcea, Rada",
  editor = "Che, Wanxiang  and
      Nabende, Joyce  and
      Shutova, Ekaterina  and
      Pilehvar, Mohammad Taher",
  booktitle = "Findings of the Association for Computational Linguistics: ACL 2025",
  month = jul,
  year = "2025",
  address = "Vienna, Austria",
  publisher = "Association for Computational Linguistics",
  url = "https://aclanthology.org/2025.findings-acl.707/",
  doi = "10.18653/v1/2025.findings-acl.707",
  pages = "13750--13770",
  ISBN = "979-8-89176-256-5"
}

@article{kadavath2022language,
  title={Language Models (Mostly) Know What They Know},
  author={Kadavath, Saurav and Conerly, Tom and Askell, Amanda and Henighan, Tom and Drain, Dawn and Perez, Ethan and Schiefer, Nicholas and Hatfield-Dodds, Zac and others},
  journal={arXiv preprint arXiv:2207.05221},
  year={2022},
  url={https://arxiv.org/abs/2207.05221}
}

@inproceedings{manakul2023selfcheckgpt,
  title = "{S}elf{C}heck{GPT}: Zero-Resource Black-Box Hallucination Detection for Generative Large Language Models",
  author = "Manakul, Potsawee  and
      Liusie, Adian  and
      Gales, Mark",
  editor = "Bouamor, Houda  and
      Pino, Juan  and
      Bali, Kalika",
  booktitle = "Proceedings of the 2023 Conference on Empirical Methods in Natural Language Processing",
  month = dec,
  year = "2023",
  address = "Singapore",
  publisher = "Association for Computational Linguistics",
  url = "https://aclanthology.org/2023.emnlp-main.557/",
  doi = "10.18653/v1/2023.emnlp-main.557",
  pages = "9004--9017"
}

@article{farquhar2024detecting,
  title={Detecting hallucinations in large language models using semantic entropy},
  author={Farquhar, Sebastian and Kossen, Jannik and Kuhn, Lorenz and Gal, Yarin},
  journal={Nature},
  volume={630},
  number={8017},
  pages={625--630},
  year={2024},
  publisher={Nature Publishing Group},
  doi={10.1038/s41586-024-07421-0}
}

@article{phillips2025geometric,
  title={Geometric Uncertainty for Detecting and Correcting Hallucinations in LLMs},
  author={Phillips, Edward and Wu, Sean and Molaei, Soheila and Belgrave, Danielle and Thakur, Anshul and Clifton, David},
  journal={arXiv preprint arXiv:2509.13813},
  year={2025},
  url={https://arxiv.org/abs/2509.13813}
}

@inproceedings{zhang2025mitigating,
  title={Mitigating Interviewer Bias in Multimodal Depression Detection: An Approach with Adversarial Learning and Contextual Positional Encoding},
  author={Zhang, Enshi and Poellabauer, Christian},
  booktitle={Findings of the Association for Computational Linguistics: EMNLP 2025},
  pages={12169--12188},
  year={2025},
  publisher={Association for Computational Linguistics},
  url={https://aclanthology.org/2025.findings-emnlp.650}
}

@inproceedings{zhang2025speecht,
    title = "{S}peech{T}-{RAG}: Reliable Depression Detection in {LLM}s with Retrieval-Augmented Generation Using Speech Timing Information",
    author = "Zhang, Xiangyu  and      Liu, Hexin  and      Zhang, Qiquan  and      Ahmed, Beena  and      Epps, Julien",
    editor = "Che, Wanxiang  and      Nabende, Joyce  and      Shutova, Ekaterina  and      Pilehvar, Mohammad Taher",
    booktitle = "Findings of the Association for Computational Linguistics: ACL 2025",
    month = jul,
    year = "2025",
    address = "Vienna, Austria",
    publisher = "Association for Computational Linguistics",
    url = "https://aclanthology.org/2025.findings-acl.521/",
    doi = "10.18653/v1/2025.findings-acl.521",
    pages = "10019--10030",
    ISBN = "979-8-89176-256-5",
}

@inproceedings{zhang2025explainable,
  title={Explainable Depression Detection in Clinical Interviews with Personalized Retrieval-Augmented Generation},
  author={Zhang, Linhai and Gao, Ziyang and Zhou, Deyu and He, Yulan},
  booktitle={Findings of the Association for Computational Linguistics: ACL 2025},
  pages={9927--9944},
  year={2025},
  publisher={Association for Computational Linguistics},
  url={https://aclanthology.org/2025.findings-acl.517}
}

@article{ringeval2019avec,
author = {Ringeval, Fabien and Schuller, Bj\"{o}rn and Valstar, Michel and Cummins, Nicholas and Cowie, Roddy and Tavabi, Leili and Schmitt, Maximilian and Alisamir, Sina and Amiriparian, Shahin and Messner, Eva-Maria and Song, Siyang and Liu, Shuo and Zhao, Ziping and Mallol-Ragolta, Adria and Ren, Zhao and Soleymani, Mohammad and Pantic, Maja},
title = {AVEC 2019 Workshop and Challenge: State-of-Mind, Detecting Depression with AI, and Cross-Cultural Affect Recognition},
year = {2019},
isbn = {9781450369138},
publisher = {Association for Computing Machinery},
address = {New York, NY, USA},
url = {https://doi.org/10.1145/3347320.3357688},
doi = {10.1145/3347320.3357688},
booktitle = {Proceedings of the 9th International on Audio/Visual Emotion Challenge and Workshop},
pages = {3–12},
numpages = {10},
location = {Nice, France},
series = {AVEC '19}
}

@inproceedings{nguyen2025beyond,
    title = "Beyond Semantic Entropy: Boosting {LLM} Uncertainty Quantification with Pairwise Semantic Similarity",
    author = "Nguyen, Dang  and
      Payani, Ali  and
      Mirzasoleiman, Baharan",
    editor = "Che, Wanxiang  and
      Nabende, Joyce  and
      Shutova, Ekaterina  and
      Pilehvar, Mohammad Taher",
    booktitle = "Findings of the Association for Computational Linguistics: ACL 2025",
    month = jul,
    year = "2025",
    address = "Vienna, Austria",
    publisher = "Association for Computational Linguistics",
    url = "https://aclanthology.org/2025.findings-acl.234/",
    doi = "10.18653/v1/2025.findings-acl.234",
    pages = "4530--4540",
    ISBN = "979-8-89176-256-5"
}

@article{PennyDimri2025Measuring,
  author    = {Penny-Dimri, Jahan C. and Bachmann, Magdalena and Cooke, William R. and Mathewlynn, Sam and Dockree, Samuel and Tolladay, John and Kossen, Jannik and Li, Lin and Gal, Yarin and Davis Jones, Gabriel},
  title     = {Measuring large language model uncertainty in women's health using semantic entropy and perplexity: a comparative study},
  journal   = {The Lancet Obstetrics, Gynaecology, \& Women’s Health},
  year      = {2025},
  volume    = {1},
  number    = {1},
  pages     = {e47--e56},
  month     = sep,
  publisher = {Elsevier},
  doi       = {10.1016/j.lanogw.2025.100005},
  url       = {https://doi.org/10.1016/j.lanogw.2025.100005},
  issn      = {3050-5038}
}

@article{Asgari2025framework,
  author  = {Asgari, Elham and Monta{\~n}a-Brown, Nina and Dubois, Magda and Khalil, Saleh and Balloch, Jasmine and Yeung, Joshua Au and Pimenta, Dominic},
  title   = {A framework to assess clinical safety and hallucination rates of LLMs for medical text summarisation},
  journal = {npj Digital Medicine},
  year    = {2025},
  volume  = {8},
  number  = {1},
  pages   = {274},
  doi     = {10.1038/s41746-025-01670-7},
  url     = {https://doi.org/10.1038/s41746-025-01670-7},
  issn    = {2398-6352}
}

@article{SARMA2026integrating,
title = {Integrating expert knowledge into large language models improves performance for psychiatric reasoning and diagnosis},
journal = {Psychiatry Research},
volume = {355},
pages = {116844},
year = {2026},
issn = {0165-1781},
doi = {https://doi.org/10.1016/j.psychres.2025.116844},
url = {https://www.sciencedirect.com/science/article/pii/S0165178125004895},
author = {Karthik V Sarma and Kaitlin E Hanss and Andrew J M Halls and Andrew Krystal and Daniel F Becker and Anne L Glowinski and Atul J Butte}
}

@article{Lee2026interpretable,
  author  = {Lee, Jae-Joong and Han, Jihoon and Woo, Choong-Wan},
  title   = {Interpretable depression assessment using a large language model},
  journal = {PLOS Digital Health},
  year    = {2026},
  volume  = {5},
  number  = {2},
  pages   = {e0001205},
  month   = feb,
  doi     = {10.1371/journal.pdig.0001205},
  url     = {https://doi.org/10.1371/journal.pdig.0001205},
  issn    = {2767-3170},
  pmid    = {41662257},
  pmcid   = {PMC12885269}
}

@misc{moon2025depressllm,
      title={DepressLLM: Interpretable domain-adapted language model for depression detection from real-world narratives}, 
      author={Sehwan Moon and Aram Lee and Jeong Eun Kim and Hee-Ju Kang and Il-Seon Shin and Sung-Wan Kim and Jae-Min Kim and Min Jhon and Ju-Wan Kim},
      year={2025},
      eprint={2508.08591},
      archivePrefix={arXiv},
      primaryClass={cs.CL},
      url={https://arxiv.org/abs/2508.08591}, 
}

@article{jinpeng2025psycollm,
  author={Hu, Jinpeng and Dong, Tengteng and Luo, Gang and Ma, Hui and Zou, Peng and Sun, Xiao and Guo, Dan and Yang, Xun and Wang, Meng},
  journal={IEEE Transactions on Computational Social Systems}, 
  title={PsycoLLM: Enhancing LLM for Psychological Understanding and Evaluation}, 
  year={2025},
  volume={12},
  number={2},
  pages={539-551},
  doi={10.1109/TCSS.2024.3497725}}

@article{Sadeghi2024harnessing,
  author  = {Sadeghi, Misha and Richer, Robert and Egger, Bernhard and Schindler-Gmelch, Lena and Rupp, Lydia Helene and Rahimi, Farnaz and Berking, Matthias and Eskofier, Bjoern M.},
  title   = {Harnessing multimodal approaches for depression detection using large language models and facial expressions},
  journal = {npj Mental Health Research},
  year    = {2024},
  volume  = {3},
  number  = {1},
  pages   = {66},
  doi     = {10.1038/s44184-024-00112-8},
  url     = {https://doi.org/10.1038/s44184-024-00112-8},
  issn    = {2731-4251}
}

@article{Li2025large,
   title={Large language models for depression recognition in spoken language integrating psychological knowledge},
   volume={7},
   ISSN={2624-9898},
   url={http://dx.doi.org/10.3389/fcomp.2025.1629725},
   DOI={10.3389/fcomp.2025.1629725},
   journal={Frontiers in Computer Science},
   publisher={Frontiers Media SA},
   author={Li, Yupei and Shao, Shuaijie and Milling, Manuel and Schuller, Björn W.},
   year={2025},
   month=Aug }

@inproceedings{bi2025magi,
    title = "{MAGI}: Multi-Agent Guided Interview for Psychiatric Assessment",
    author = "Bi, Guanqun  and
      Chen, Zhuang  and
      Liu, Zhoufu  and
      Wang, Hongkai  and
      Xiao, Xiyao  and
      Xie, Yuqiang  and
      Zhang, Wen  and
      Huang, Yongkang  and
      Chen, Yuxuan  and
      Peng, Libiao  and
      Huang, Minlie",
    editor = "Che, Wanxiang  and
      Nabende, Joyce  and
      Shutova, Ekaterina  and
      Pilehvar, Mohammad Taher",
    booktitle = "Findings of the Association for Computational Linguistics: ACL 2025",
    month = jul,
    year = "2025",
    address = "Vienna, Austria",
    publisher = "Association for Computational Linguistics",
    url = "https://aclanthology.org/2025.findings-acl.1278/",
    doi = "10.18653/v1/2025.findings-acl.1278",
    pages = "24898--24921",
    ISBN = "979-8-89176-256-5"
}

@article{Wu2026wisemind,
  author  = {Wu, Yuqi and Wan, Guangya and Li, Jingjing and Zhao, Shengming and Ma, Lingfeng and Ye, Tianyi and Zhang, Mike and Pop, Ion and Zhang, Yanbo and Chen, Jie},
  title   = {WiseMind: a knowledge-guided multi-agent framework for accurate and empathetic psychiatric diagnosis},
  journal = {npj Digital Medicine},
  year    = {2026},
  doi     = {10.1038/s41746-026-02559-9},
  url     = {https://doi.org/10.1038/s41746-026-02559-9},
  issn    = {2398-6352}
}

@misc{huang2020clinicalbert,
      title={ClinicalBERT: Modeling Clinical Notes and Predicting Hospital Readmission}, 
      author={Kexin Huang and Jaan Altosaar and Rajesh Ranganath},
      year={2020},
      eprint={1904.05342},
      archivePrefix={arXiv},
      primaryClass={cs.CL},
      url={https://arxiv.org/abs/1904.05342}, 
}

@misc{labrak2024biomistral,
      title={BioMistral: A Collection of Open-Source Pretrained Large Language Models for Medical Domains}, 
      author={Yanis Labrak and Adrien Bazoge and Emmanuel Morin and Pierre-Antoine Gourraud and Mickael Rouvier and Richard Dufour},
      year={2024},
      eprint={2402.10373},
      archivePrefix={arXiv},
      primaryClass={cs.CL},
      url={https://arxiv.org/abs/2402.10373}, 
}

@misc{chen2023meditron,
      title={MEDITRON-70B: Scaling Medical Pretraining for Large Language Models}, 
      author={Zeming Chen and Alejandro Hernández Cano and Angelika Romanou and Antoine Bonnet and Kyle Matoba and Francesco Salvi and Matteo Pagliardini and Simin Fan and Andreas Köpf and Amirkeivan Mohtashami and Alexandre Sallinen and Alireza Sakhaeirad and Vinitra Swamy and Igor Krawczuk and Deniz Bayazit and Axel Marmet and Syrielle Montariol and Mary-Anne Hartley and Martin Jaggi and Antoine Bosselut},
      year={2023},
      eprint={2311.16079},
      archivePrefix={arXiv},
      primaryClass={cs.CL},
      url={https://arxiv.org/abs/2311.16079}, 
}

\vfill

\end{document}